\documentclass[11pt]{article}

\usepackage[preprint]{acl}

\usepackage{times}
\usepackage{latexsym}

\usepackage[T1]{fontenc}
\usepackage[utf8]{inputenc}

\usepackage{microtype}

\usepackage{inconsolata}

\usepackage{graphicx}

\title{ReaORE: Reasoning-Guided Progressive Open Relation Extraction Empowered by Large Reasoning Models}
\author{
Xin Lin$^{1,2}$\Thanks{Equal contribution.} \quad
Liang Zhang$^{2}$\footnotemark[1] \quad
Guoqi Ma$^{2}$ \quad
Hongyao Tu$^{2}$ \quad
Jinsong Su$^{1,2}$\Thanks{Corresponding author.} \\
$^{1}$National Institute for Data Science in Health and Medicine, Xiamen University, China \\
$^{2}$School of Informatics, Xiamen University \\
\texttt{\{linxin1,lzhang\}@stu.xmu.edu.cn, jssu@xmu.edu.cn}
}

\usepackage{svg}
\usepackage{amsmath}
\usepackage{amssymb}
\usepackage{booktabs}
\usepackage{multirow}
\usepackage{colortbl}

\begin{document}
\maketitle
\begin{abstract}
% 1.任务定义：开放关系抽取（OpenRE）需要模型来抽取非结构化文本中头部和尾部实体之间未见的关系，以用于实际应用。
Open Relation Extraction (OpenRE) requires model to extract unseen relation between head and tail entities from unstructured text for real-world applications.
% 2.gap：OpenRE 的核心挑战在于，模型需要在未见过目标关系的情况下，仍能准确泛化。
The core challenge of OpenRE lies in achieving reliable generalization to unseen relation types.
% 3.当前的 OpenRE 方法要么采用聚类技术，但其无法生成关系标签，泛化能力较差；要么方法依赖于通过大型语言模型 (LLM) 直接生成关系标签，但这方法不足以区分容易混淆的关系。
Current OpenRE approaches either employ clustering techniques, which cannot generate relation labels and suffer from poor generalization, or rely on direct relation label generation via Large Language Models (LLMs), which lack sufficient discriminative capacity to distinguish easily confused relations.
% 4.为此，我们提出 ReaORE，这是一个通过由粗到精的关系推理来执行关系抽取的框架。
To address these limitations, we propose \textbf{Rea}soning-guided progressive \textbf{O}pen\textbf{RE} (ReaORE), a framework for performing relation extraction through coarse-to-fine relation reasoning. 
% 5.具体来说，ReaORE 由两个关键阶段组成：（i）关系过滤，它通过多方面推理来理解关系和实例，从而产生初始关系集，并通过基于嵌入的相似性进一步补充和过滤关系，以确保包含目标关系；（ii）关系预测，旨在通过细粒度的比较推理从上述集合中预测目标关系，以便更好地区分容易混淆的关系。
Specifically, ReaORE consists of two key stages: (i) relation filtering, which reasons over multiple aspects to understand relations and instances, yielding an initial relation set, and further supplements and filters relations via embedding-based similarity to ensure the target relation is included; (ii) relation prediction, which aims to predict the target relations from the above set via fine-grained comparative reasoning to better distinguish easily confused relations.
% We further construct a training dataset containing explicit reasoning chains to enhance the instruction-following capability of Large Reasoning Models~(LRMs) for both matching and contrastive reasoning.
% 5.实验结果
Extensive experiments on two widely used OpenRE datasets demonstrate that ReaORE outperforms existing baselines.\footnote{Our code will be released upon acceptance.}
\end{abstract}

\section{Introduction}
\begin{figure}[t]
    \centering
    % 注意：这里使用 \includesvg 替代 \includegraphics
    \includegraphics[width=1.0\linewidth]{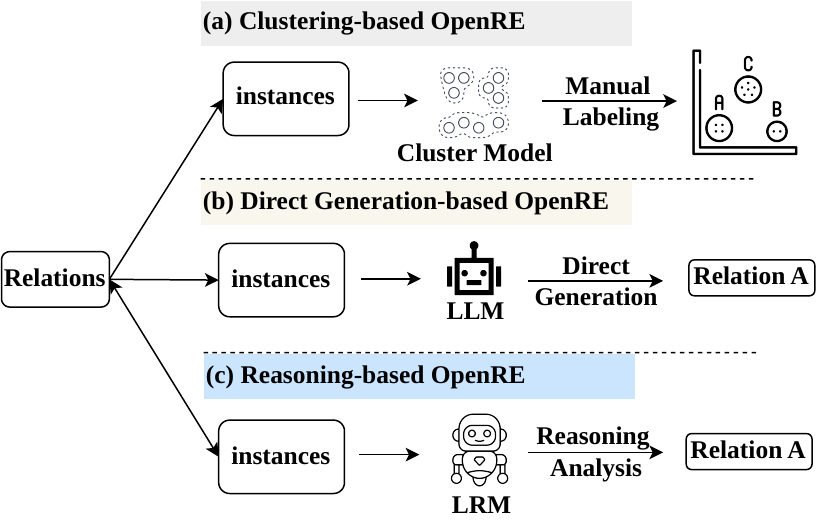} 
    \caption{Comparison of open relation extraction methods.}
    \label{fig:intro}
\end{figure}
% re
% 关系抽取（RE）旨在识别非结构化文本中头实体与尾实体之间潜在的关系。作为一项基础的信息抽取任务，RE提供了必要的关系事实，为知识库构建~\cite{ji2011knowledge}和问答系统~\cite{yu2017improved}等各类下游应用提供支持。
Relation Extraction (RE) aims to identify potential relations between head and tail entities within unstructured text~\cite{diaz2025survey}. As a fundamental information extraction task, it provides essential relational fact information that support various downstream applications, such as knowledge base construction~\cite{ji2011knowledge} and question answering systems~\cite{yu2017improved}. 
% openre及其挑战
% 传统关系抽取本质上局限于预定义的关系类型，无法泛化到现实世界中不断涌现的关系。为了解决这个问题，开放关系抽取（OpenRE）~\cite{zhao2023actively,LLMOERF_tu2025llm} 得到了广泛的研究，旨在抽取训练过程中未曾见过的全新关系类型。
Traditional RE is inherently limited to predefined relation types, failing to generalize to the ever-emerging relations in real-world scenarios. To address this, Open Relation Extraction (OpenRE)~\cite{zhao2023actively,LLMOERF_tu2025llm} has been widely studied, aiming to extract novel relation types unseen during training.
% 然而，由于目标关系类型在训练阶段不可见，模型缺乏足够的判别特征来刻画这些关系，导致对于未见关系的泛化能力不佳。
Nevertheless, since target relation types are unseen during training, models lack sufficient discriminative features to characterize them, resulting in poor generalization to unseen relations.

% openre定义？——To address this limitation, researchers have turned to Open Relation Extraction (OpenRE), which is not confined to a predefined set of relations and can dynamically discover new ones, making it more practical for real-world applications.  //The goal of open relation extraction (OpenRE) is to develop an RE model that can generalize to new relations not encountered during training. 为了克服这一局限性，研究人员转向了开放关系抽取（OpenRE）。
% OpenRE 不局限于预定义的关系集，能够动态地发现新的关系，使其更适用于实际应用。// 开放关系抽取（OpenRE）的目标是开发一种关系抽取模型，使其能够泛化到训练过程中未遇到的新关系。

% 现有工作分类
% 现有 OpenRE 方法大致可分为两类：基于聚类的方法和基于 LLM 直接生成关系标签的方法。
Existing OpenRE methods can be broadly categorized into two categories: clustering-based methods and LLM-based direct relation label generation methods.
% 基于聚类
% 前者通常将实例及一些重要的相关信息编码成潜在空间内的嵌入。随后，这些嵌入向量会依据其相对距离进行聚类，其中每一个聚类代表一种潜在的未知关系。
The former typically encodes instances and some important related information into embeddings within a latent space~\cite{hu2020selfore,RoCORE_zhao2021relation,zhao2023actively,mixore_wang2025towards,orellm_wang2024phrases}. These embeddings are subsequently clustered based on their relative distances, with each cluster representing a potentially unseen relation, as illustrated in Figure~\ref{fig:intro}(a). 
% 然而，为了给每个聚类赋予语义标签需要耗费大量人力进行后处理，本质上执行的是关系聚类而非关系抽取，且此类方法对训练要求高导致泛化能力差。
However, assigning semantic labels to each cluster requires labor-intensive post-processing, making these methods essentially perform relation clustering rather than relation extraction. Moreover, their heavy reliance on training data leads to poor generalization.
% 基于大语言模型生成
% 鉴于强大的大型语言模型（LLMs）在各类自然语言处理任务中展现出的卓越能力，它们在为 OpenRE 提供先验知识与语义理解方面，蕴含着巨大的潜力。
Given the powerful LLMs' capabilities across diverse natural language processing tasks~\cite{xu2024large}, they hold considerable potential to provide OpenRE with both prior knowledge and semantic understanding. 
% 更重要的是，与传统的聚类方法不同，LLM 的生成特性使其能够以自然语言形式直接预测关系。
More importantly, unlike traditional clustering-based methods, the generative nature of LLMs enables them to directly predict relations in natural language forms~\cite{LLMOERF_tu2025llm}, as illustrated in Figure~\ref{fig:intro}(b).
% 然而，直接生成关系的方式使模型无法从多角度综合理解实例与关系，导致模型缺乏足够的判别能力来区分易混淆关系。
However, direct relation generation prevents the model from comprehensively understanding instances and relations from multiple perspectives, resulting in insufficient discriminative capacity to distinguish easily confused relations.

% 推理对于openre文本关系对齐的必要性：
% 大型推理模型（LRMs）~\cite{dsr1,o1,qwq} 的核心特征在于能够生成显式且详尽的长思维链，这赋予了模型更强的分析、理解与推理能力，已在数学推理、代码生成等多个复杂任务中得到充分验证。
Large Reasoning Models (LRMs)~\cite{dsr1,o1,qwq} are characterized by their ability to generate explicit and extensive long reasoning chains, which endow models with stronger capacities for analysis, comprehension, and inference, as demonstrated across complex tasks such as mathematical reasoning and code generation.~\cite{huang2025math,pelaccia2025diagnostic,wang2025medagent}. 
% 得益于这种深度理解能力，LRMs 能够渐进式地对实例和关系进行由粗到细的深层分析，从而更准确地理解关系语义与实例特征。
Leveraging this deep understanding capability, LRMs can progressively analyze instances and relations from coarse to fine, achieving more accurate comprehension of relation semantics and instance characteristics. 
% 更为重要的是，推理能力本质上是一种通用能力，模型通过 reasoning 真正理解关系语义，因此面对任何未见关系都能做出有据可依的判断，这与 OpenRE 对泛化能力的核心要求高度契合。
More critically, reasoning represents a generalizable capability: by genuinely understanding relation semantics, LRMs can produce well-grounded predictions for unseen relation types, which precisely addresses the core generalization requirement of OpenRE. 
% 此外，即使面对易混淆关系，reasoning 也能通过深度分析显式给出判断依据，从而做出可靠预测。
Furthermore, even for easily confused relations, reasoning enables the model to explicitly derive judgment evidence through in-depth analysis, yielding reliable predictions. 
% 因此，将 reasoning 引入 OpenRE 是极具潜力的研究方向。
Therefore, incorporating reasoning into OpenRE represents a highly promising research direction.

% In this paper,我们提出了一种基于推理的开放式关系抽取框架（ReaORE），它通过显式且基于证据的关系分析，执行由粗到细的关系抽取。在ReaORE中，第一阶段是粗粒度关系覆盖，它通过对全局语义和实体类型的推理，识别与输入实例兼容的关系，并通过基于嵌入的相似性进一步补充被忽略的关系，以确保全面覆盖。第二阶段是细粒度关系预测，它从上述关系集中选择高置信度的关系进行细粒度对比推理，显式地生成排他性判断证据，以准确区分容易混淆的关系并确定最终预测结果。
In this paper, we propose a reasoning-guided progressive OpenRE framework (ReaORE) that performs coarse-to-fine relation extraction through explicit relation reasoning. In ReaORE, the first stage is relation filtering, which finds relations compatible with the input instance by coarse-grained reasoning about global semantics and entity types, and further supplements and filters them by embedding-based similarity. The second stage is relation prediction, which performs fine-grained contrastive reasoning, explicitly generating exclusive judgment evidence to accurately distinguish easily confused relations and determine the final prediction.

% 介绍ReaORE
% 为了应对这些挑战，我们提出了 ReaORE——一个基于大型推理模型（LRM）的 OpenRE 框架，通过审慎的渐进式关系推理实现可解释的关系抽取，摆脱对实例聚类或直接生成的依赖。
% To address these challenges, we propose ReaORE, an LRM-based OpenRE framework that achieves interpretable relation extraction through deliberate progressive relational reasoning, moving beyond instance clustering and direct generation.
% 在 ReaORE 中，输入实例首先执行匹配推理，从全局语义和头尾实体类型两个维度评估其与各关系标签的匹配程度，得出初步候选关系集合。随后执行关系召回，基于嵌入相似度检索与查询语义相近的关系补充至候选集合，并从中选出 top-k个候选，从而优化关系覆盖、避免错误传播。最终，这些候选关系被送入对比推理模块，进行细粒度的对比推理，生成排他性判断依据，准确区分语义相近的关系。
% In ReaORE, input instance first undergoes matching-based reasoning to evaluate its compatibility with various relation labels in terms of global semantics and entity types, yielding an initial candidate set. Relation recall is then performed, retrieving relations semantically similar to the query via embedding-based similarity and supplementing them into the candidate set, from which the top-k candidates are selected to improve relation coverage and mitigate error propagation. Finally, these candidates are passed into the contrastive reasoning module, where fine-grained contrastive reasoning generates mutually exclusive evidence to accurately distinguish semantically similar relations.
% 此外，我们利用先进的大型语言模型（LLM）来构建专为关系推理模型量身定制的高质量训练数据集。
% Additionally, we utilize an advanced LLM to construct high-quality training datasets tailored for the relation reasoning model.

In summary, our main contributions are as follows:
\begin{itemize}
    % 我们提出 ReaORE，一个基于推理的 OpenRE 框架，通过推理增强模型对未见新关系的泛化能力。
    \item We propose ReaORE, a reasoning-based progressive OpenRE framework that enhances the model's generalization capability to unseen relation types through reasoning.
    % 我们在关系预测阶段采用细粒度对比推理策略，生成排他性证据，显式给出选择某一关系而非其他关系的理由，有效消除易混淆关系间的歧义。
    \item We introduce a fine-grained contrastive reasoning strategy to effectively distinguish easily confused relations, thereby achieving better relation generalization.
    % 在两个广泛使用的 OpenRE 基准数据集上的大量实验表明，ReaORE 显著优于所有现有基线方法，为推理赋能开放式关系抽取提供了可行方案。
    \item Extensive experiments on two widely used OpenRE benchmark datasets demonstrate that ReaORE significantly outperforms existing baseline methods, providing a viable solution for empowering OpenRE with deliberate reasoning.
\end{itemize}
\begin{figure*}[!t]
    \centering
    % 注意：这里使用 \includesvg 替代 \includegraphics
    \includegraphics[width=1.0\linewidth]{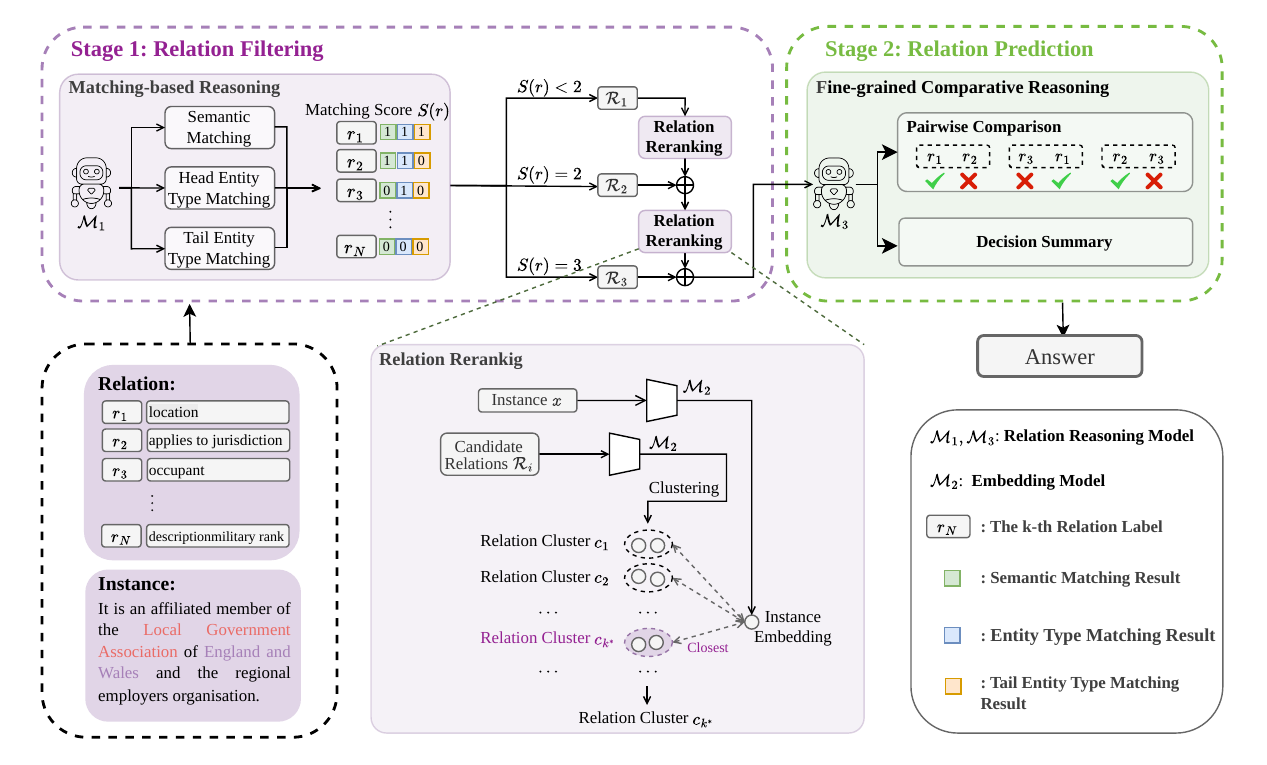} 
    % 匹配推理模型和对比推理模型由本文构造的长思维链训练数据集分别经过微调得到，而重排模型是一个不需要训练的embedding model。
    \caption{Overview of ReaORE Framework. ReaORE follows a coarse-to-fine process: it first filters unseen relations through matching-based reasoning $\mathcal{M}_{1}$ and embedding model $\mathcal{M}_{2}$, and then predicts the final relation via fine-grained contrastive reasoning $\mathcal{M}_{3}$. $\mathcal{M}_{1}$ and $\mathcal{M}_{3}$ are fine-tuned with long-reasoning-chain data, whereas $\mathcal{M}_{2}$ is training-free.}
    \label{fig:main}
\end{figure*}

\section{Task Formulation}
Before presenting our framework, we first introduce the preliminary notations.
Let $\mathcal{R}_{k}$ denote the set of known relations observed during training and $\mathcal{R}_{u}$ denote the set of unseen relations used for evaluation, where $\mathcal{R}_{k} \cap \mathcal{R}_{u} = \emptyset$.
During training, the model can access labeled instances associated with relations in $\mathcal{R}_{k}$.
At test time, given an unlabeled dataset $\mathcal{D}=\{x_i\}_{i=1}^{N}$, each instance $x_i=\langle s_i,h_i,t_i\rangle$ consists of a sentence $s_i$, a head entity $h_i$, and a tail entity $t_i$.
The goal is to predict a relation label $\hat{r}_i \in \mathcal{R}_{u}$ that best describes the semantic relation expressed between $h_i$ and $t_i$ in $s_i$.

For each relation extraction, we do not provide additional information such as instance descriptions or supplementary entity information. ReaORE leverages the prior knowledge and semantic understanding of LRM to infer relation semantics and head and tail entity constraints from the input context.
This formulation reflects a real-world OpenRE setting where new, unknown relations emerge continuously, but there are no labeled training instances provided for these unseen relations, and no supplementary knowledge beyond the original instance text to aid analysis.
Since clustering-based methods and direct relation label generation methods often suffer from limited generalization and unreliable predictions, ReaORE first derives a compact candidate set $\mathcal{R}_c$ through matching-based reasoning and embedding-based relation reranking, and then performs fine-grained contrastive reasoning over $\mathcal{R}_c$ to determine the final prediction.

\section{Methodology}
% 在本节中，我们介绍了所提出的 ReaORE，这是一个用于OpenRE（开放域关系抽取）文本与未见关系标签对齐的两阶段推理框架。我们的方法融合了用于获取初步对齐线索的匹配推理，以及用于区分语义相似关系的细粒度对比推理。图~\ref{fig:main} 概述了该框架的整体结构。
In this section, we present ReaORE, a reasoning-guided progressive framework for OpenRE that extracts unseen relations through coarse-to-fine relation reasoning. Rather than directly generating a relation label or clustering instances, ReaORE decomposes prediction into two stages: relation filtering and relation prediction. Relation filtering first identifies relations compatible with the input instance through coarse-grained reasoning and then refines $\mathcal{R}_c$ via embedding-based reranking to improve relation coverage. Relation prediction further applies fine-grained contrastive reasoning over $\mathcal{R}_c$, producing explicit judgment evidence for the final decision. Figure~\ref{fig:main} provides an overview of the framework.

\subsection{Relation Filtering}
Relation filtering aims to preliminarily construct $\mathcal{R}_c$ over unseen relation types, making it compact while preserving high coverage.
It contains two complementary steps: matching-based reasoning first provides coarse-grained binary judgments, and relation reranking then uses embedding-space similarity to recover plausible relations that may be missed by the reasoning model.

\textbf{Matching-Based Reasoning. }
For each input instance $x_i=\langle s_i,h_i,t_i\rangle$, the matching-based inference model $\mathcal{M}_{1}$ first performs query analysis on the sentence $s_i$ to better understand its meaning and then analyzes the head and tail entity types within the given context.
This step makes the input evidence more explicit, ensuring that subsequent relation matching is based on sentence facts and entity types, not just surface label similarity.

\begin{figure*}[!t]
    \centering
    \begin{minipage}[t]{0.49\linewidth}
        \centering
        \includegraphics[width=\linewidth]{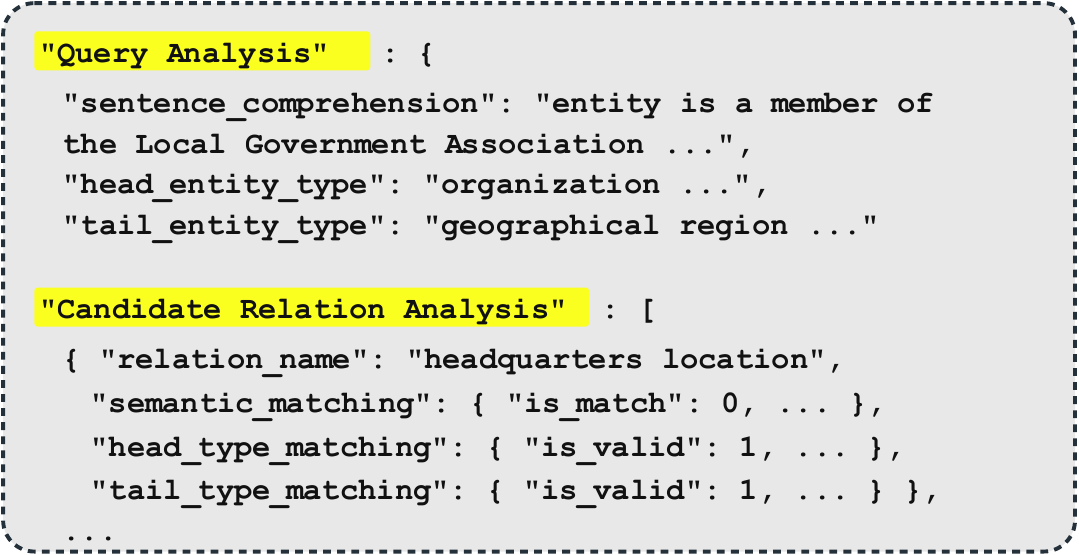}
        \vspace{-0.5em}
        \centerline{\small (a) Matching-based Reasoning Record.}
    \end{minipage}
    \hfill
    \begin{minipage}[t]{0.49\linewidth}
        \centering
        \includegraphics[width=\linewidth]{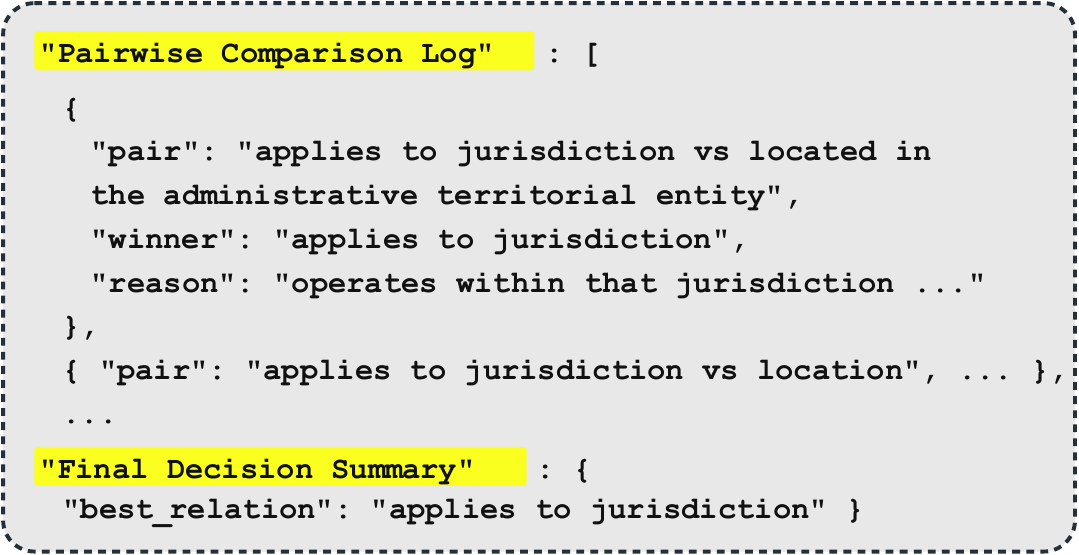}
        \vspace{-0.5em}
        \centerline{\small (b) Comparative Reasoning Record.}
    \end{minipage}
    \caption{Examples of the structured reasoning records produced by ReaORE. (a) The matching-based reasoning record analyzes the query and evaluates relations from semantic, head-entity type, and tail-entity type perspectives. (b) The comparative reasoning record summarizes pairwise comparisons over $\mathcal{R}_c$ and the final decision.}
    \label{fig:reasoning_record_example}
\end{figure*}

After query analysis, $\mathcal{M}_{1}$ continues to generate reasoning that performs pattern matching for each relation $r$.
The model first interprets the natural-language meaning of $r$ and infers its expected head- and tail-side argument constraints.
It then judges whether $r$ matches $x_i$ from three aspects.
\textbf{Semantic matching} determines whether the relation meaning is explicitly supported by the facts in $s_i$; the model must identify the key evidence supporting or contradicting $r$, or state what information is missing.
\textbf{Head-entity type matching} checks whether $h_i$ satisfies the head-side argument constraint of $r$, while \textbf{tail-entity type matching} checks whether $t_i$ satisfies the tail-side argument constraint.
Each matching result is represented by a Boolean value, where 1 denotes a match and 0 denotes a mismatch, and every judgment is accompanied by an explanatory rationale.
Figure~\ref{fig:reasoning_record_example}(a) gives an abbreviated example of this matching-based reasoning record.

The total matching score for relation $r$ is computed as $S(r)=S_s(r)+S_h(r)+S_t(r)$, where $S_s(r), S_h(r), S_t(r)$ $\in$ $\{0,1\}$ denote the binary judgments for semantic matching, head-entity type matching, and tail-entity type matching, respectively.
A higher $S(r)$ means that more matching aspects support relation $r$ being expressed by $x_i$.
Based on these multi-aspect judgments, $\mathcal{M}_{1}$ produces the initial relation tiers in the relation filtering stage: $\mathcal{R}_{1}=\{r\mid S(r)<2\}$, $\mathcal{R}_{2}=\{r\mid S(r)=2\}$, and $\mathcal{R}_{3}=\{r\mid S(r)=3\}$.

\textbf{Relation Reranking. }
After $\mathcal{M}_{1}$ produces the score-based relation tiers, some plausible relations may still be placed in lower tiers because the sentence expresses the relation implicitly or the matching evidence is incomplete.
To improve target-relation coverage, ReaORE applies the embedding model $\mathcal{M}_{2}$ to supplement the tiers through embedding-based similarity.
As shown in Figure~\ref{fig:main}, $\mathcal{R}_{i}$ denotes one of the score-based tiers produced by $\mathcal{M}_{1}$, where $i \in \{1,2,3\}$.
For a source tier $\mathcal{S}$, $\mathcal{M}_{2}$ encodes the input instance $x_i$ and all relations in $\mathcal{S}$.
It partitions $\mathcal{S}$ into relation clusters $\mathcal{C}(\mathcal{S})=\{C_{1},\ldots,C_{K}\}$ based on relation embeddings, where each $C_j\subseteq\mathcal{S}$ is a set of relation labels.
Let $\mathbf{m}_{j}$ denote the centroid embedding of cluster $C_j$, computed by mean pooling the embeddings of relations in $C_j$.
The cluster most similar to $x_i$ is selected by:
\begin{equation}
    k^*(\mathcal{S}) = \arg\max_j \cos(\mathbf{m}_{j}, \mathbf{e}_{x_i}),
\end{equation}
where $\mathbf{e}_{x_i}$ is the embedding of the input instance.
We denote the selected relation cluster as $C^*(\mathcal{S})=C_{k^*(\mathcal{S})}$.
ReaORE applies this operation progressively:
\begin{equation}
    \widetilde{\mathcal{R}}_{2}
    = \mathcal{R}_{2} \cup C^*(\mathcal{R}_{1}),
    \quad
    \widetilde{\mathcal{R}}_{3}
    = \mathcal{R}_{3} \cup C^*(\widetilde{\mathcal{R}}_{2}).
\end{equation}
In Figure~\ref{fig:main}, relation reranking is applied twice.
First, $\mathcal{R}_{1}$ is reranked by $\mathcal{M}_{2}$, and the selected cluster $C^*(\mathcal{R}_{1})$ is merged with $\mathcal{R}_{2}$ to obtain $\widetilde{\mathcal{R}}_{2}$.
Second, $\widetilde{\mathcal{R}}_{2}$ is reranked in the same way, and the selected cluster $C^*(\widetilde{\mathcal{R}}_{2})$ is merged with $\mathcal{R}_{3}$ to obtain $\widetilde{\mathcal{R}}_{3}$.
ReaORE then ranks the relations in $\widetilde{\mathcal{R}}_{3}$ by their embedding similarity to $x_i$ and keeps the top-$\xi$ relations as $\mathcal{R}_{c}$ for $\mathcal{M}_{3}$.

\subsection{Relation Prediction}
After relation filtering, $\mathcal{R}_{c}$ is passed to the contrastive reasoning model $\mathcal{M}_{3}$ together with the input instance $x_i$.
Since these relations have already been filtered and are all relevant to the instance, they are often semantically close and difficult to distinguish, such as ``\emph{location}'' vs. ``\emph{located in the administrative territorial entity}'' or ``\emph{field of work}'' vs. ``\emph{occupation}''.
A one-step label decision is therefore unreliable, because the model may select a broadly related relation without comparing it with more precise alternatives.

As shown in Figure~\ref{fig:main}, $\mathcal{M}_{3}$ performs fine-grained comparative reasoning over $\mathcal{R}_{c}$.
For each relation pair $(r_a,r_b)$, $\mathcal{M}_{3}$ judges which relation is better supported by the sentence $s_i$ and the entity pair $(h_i,t_i)$.
Each comparison follows three criteria:
(1) Evidence Verification, which checks whether the sentence provides explicit evidence for the relation;
(2) Semantic Granularity, which prefers the more specific relation when two relations are both plausible; and
(3) Contextual Alignment, which selects the relation that best matches the event or phrase expressed in $s_i$.
The output is stored as a comparative reasoning record.
Its \textbf{Pairwise Comparison Log} records the compared pair, the winner, and the supporting reason for each comparison, while its \textbf{Final Decision Summary} reports the best relation and the decision logic.
Figure~\ref{fig:reasoning_record_example}(b) illustrates this record with abbreviated pairwise rationales, and complete traces are provided in Appendix~\ref{app:reasoning_traces}.

This stage turns relation prediction from direct label generation into explicit comparison over $\mathcal{R}_{c}$.
By requiring $\mathcal{M}_{3}$ to justify why one relation is preferred over alternatives, ReaORE better distinguishes easily confused relations and grounds the final answer in sentence evidence.
%3.4 模型训练
\subsection{Model Training}
In ReaORE, $\mathcal{M}_{1}$ and $\mathcal{M}_{3}$ denote two LRMs used for different reasoning stages.
To train their stage-specific reasoning abilities, we construct separate reasoning-chain data for matching-based reasoning and fine-grained comparative reasoning.
For $\mathcal{M}_{1}$, the training target contains query analysis and relation-wise judgments over semantic matching, head-entity type matching, and tail-entity type matching, enabling the model to construct relation tiers through multi-aspect matching.
For $\mathcal{M}_{3}$, the training target contains pairwise comparisons and a final decision summary, enabling the model to determine the final prediction through fine-grained comparative reasoning.

We construct the training data by using Gemini-2.5 Pro as the teacher model on labeled instances from the known relation set $\mathcal{R}_{k}$ only; no labeled instances from $\mathcal{R}_{u}$ are used during training.
For each stage, the teacher model is prompted to generate the corresponding stage-specific reasoning chain and answer.
The prompts define the task format and output constraints for the two reasoning stages.
During data construction, they are used to guide the teacher model in generating reasoning-chain training examples; the complete prompts are provided in Appendix~\ref{app:prompts}.
We then apply rejection sampling and retain only examples whose reasoning output leads to the ground-truth relation.
This procedure yields 1,800 high-quality examples for matching-based reasoning and 1,800 examples for fine-grained comparative reasoning.
Finally, we perform Supervised Fine-Tuning (SFT) separately on the two types of reasoning-chain training data, producing $\mathcal{M}_{1}$ and $\mathcal{M}_{3}$ used in our experiments.

\section{Experiments}
\subsection{Settings}

\textbf{Datasets. }
We conduct experiments on two widely used relation extraction datasets: FewRel~\cite{han2018fewrel} and TACRED~\cite{zhang2017position}.
To ensure fair comparison with prior OpenRE studies, we follow the relation-splitting protocol of ASCORE~\cite{zhao2023actively}.
FewRel contains 80 relation types with 700 instances for each relation; the first 40 relations are used as known relations, and the remaining 40 are treated as unseen relations.
TACRED contains 41 relation types; the first 20 relations are used as known relations, while the remaining 21 form the unseen relation set.

\textbf{Metrics. }
Following ASCORE~\cite{zhao2023actively}, we evaluate ReaORE with both clustering and classification metrics.
For clustering-oriented evaluation, we report B$^3$ precision, recall, and F$_1$~\cite{bagga1998algorithms}, V-measure homogeneity, completeness, and F$_1$~\cite{Rosenberg_Hirschberg_2007}, and Adjusted Rand Index (ARI)~\cite{hubert1985comparing}.
For relation classification, we report precision, recall, and Macro-F$_1$~\cite{opitz2019macro}.

\textbf{Baselines. }
We compare ReaORE with three SLM-based OpenRE methods and three LLM-based methods.
The SLM-based baselines include SelfORE~\cite{hu2020selfore}, RoCORE~\cite{RoCORE_zhao2021relation}, and ASCORE~\cite{zhao2023actively}.
The LLM-based baselines include vanilla Qwen3-8B~\cite{yang2025qwen3}, ORELLM~\cite{orellm_wang2024phrases}, and LLM-OREF~\cite{LLMOERF_tu2025llm}.
More details about the baselines are provided in Appendix~\ref{app:baseline_details}.

\textbf{Implementation Details. }
We use Qwen3-8B~\cite{yang2025qwen3} as the LRM backbone and standardize all LLM-based baselines to the same backbone for fair comparison.
The reasoning models are fine-tuned via LLaMA-Factory~\cite{zheng-etal-2024-llamafactory} with LoRA~\cite{hu2021loralowrankadaptationlarge} for 5 epochs.
We set the LoRA rank, scaling factor, and dropout to 64, 256, and 0.1, respectively, with a maximum sequence length of 4,096 tokens.
Optimization uses AdamW with a learning rate of $1\times10^{-4}$, a cosine scheduler, and a warmup ratio of 0.1.
BGE-M3~\cite{bge_m3} is used as the training-free embedding backbone, and $\xi$ is set to 5 by default.
We use vLLM~\cite{vllm} on 2 A100 80GB GPUs; see Appendix~\ref{app:implementation_hyperparameters}.

\begin{table*}[t]
\centering
\setlength{\tabcolsep}{0.42em}
\small
\begin{tabular}{llcccccccccc}
\toprule
\multirow{2}{*}{\textbf{Dataset}} & \multirow{2}{*}{\textbf{Method}} & \multicolumn{3}{c}{\textbf{B$^3$}} & \multicolumn{3}{c}{\textbf{V-Measure}} & \multirow{2}{*}{\textbf{ARI}} & \multicolumn{3}{c}{\textbf{Classification}} \\
\cmidrule(lr){3-5} \cmidrule(lr){6-8} \cmidrule(lr){10-12}
 &  & Prec. & Rec. & $F_{1}$ & Hom. & Comp. & $F_{1}$ &  & Prec. & Rec. & $F_{1}$ \\
\midrule
\multirow{7}{*}{\textbf{FewRel}} 
 & SelfORE~\cite{hu2020selfore} & 52.7 & 55.2 & 53.9 & 72.8 & 73.6 & 73.2 & 51.7 & 60.4 & 63.2 & 60.0 \\
 & RoCORE~\cite{RoCORE_zhao2021relation} & 80.6 & 84.3 & 82.4 & 88.3 & 89.6 & 88.9 & \textbf{80.7} & 82.7 & 86.8 & 83.7 \\
 & ASCORE~\cite{zhao2023actively} & 79.9 & 84.1 & 81.9 & 88.8 & 90.1 & \underline{89.4} & 80.1 & 83.2 & 86.2 & \underline{83.8} \\
\cmidrule{2-12}
 & Qwen3-8B~\cite{yang2025qwen3} & 62.0 & 65.9 & 63.8 & 74.6 & 78.4 & 76.4 & 53.7  & 76.4 & 70.6 & 73.0\\
 & ORELLM~\cite{orellm_wang2024phrases} & 70.8 & 72.3 & 71.5 & 80.6 & 82.7 & 81.6 & 74.7 & 85.7 & 79.4 & 79.5 \\
 & LLM-OREF~\cite{LLMOERF_tu2025llm} & 87.7 & 82.0 & \textbf{84.8} & 83.1 & 85.3 & 84.2 & 79.0 & 87.7 & 82.0 & 81.7 \\
\rowcolor{gray!10} & \textbf{Ours} & 81.3 & 83.9 & \underline{82.6} & 90.6 & 88.7 & \textbf{89.6} & \textbf{80.7} & 85.7 & 83.4 & \textbf{84.2} \\
\midrule
\multirow{7}{*}{\textbf{TACRED}} 
 & SelfORE~\cite{hu2020selfore} & 51.7 & 44.1 & 47.5 & 63.1 & 60.0 & 61.5 & 43.4 & 34.3 & 39.6 & 36.0 \\
 & RoCORE~\cite{RoCORE_zhao2021relation} & 69.6 & 68.5 & 69.0 & 78.6 & 78.6 & 78.6 & 54.7 & 59.4 & 56.3 & 56.3 \\
 & ASCORE~\cite{zhao2023actively} & 74.2 & 82.1 & \underline{77.9} & 80.7 & 85.6 & \underline{83.0} & 78.1 & 69.8 & 71.5 & 69.9 \\
\cmidrule{2-12}
 & Qwen3-8B~\cite{yang2025qwen3} & 68.8 & 67.5 & 68.1 & 74.5 & 75.8 & 75.1 & 72.5 & 63.4 & 65.1 & 64.2 \\
 & ORELLM~\cite{orellm_wang2024phrases} & 75.5 & 71.5 & 73.4 & 78.0 & 79.2 & 78.5 & \underline{79.5} & 70.5 & 69.8 & \underline{70.1} \\
 & LLM-OREF~\cite{LLMOERF_tu2025llm} & 78.8 & 76.0 & 77.3 & 80.2 & 84.3 & 82.1 & 77.8 & 69.8 & 76.9 & 68.9 \\
\rowcolor{gray!10} & \textbf{Ours} & 78.2 & 79.5 & \textbf{78.8} & 81.4 & 84.9 & \textbf{83.1} & \textbf{81.3} & 72.0 & 70.8 & \textbf{71.4} \\
\bottomrule
\end{tabular}
\caption{Performance comparison on FewRel and TACRED datasets. The best and second-best results are marked in \textbf{bold} and \underline{underlined}, respectively.}
\label{tab:main}
\end{table*}

\subsection{Main Results}
Table~\ref{tab:main} reports the main results on FewRel and TACRED.
Overall, ReaORE achieves the best or tied-best performance on the main clustering and classification metrics across both datasets, demonstrating the effectiveness of progressive reasoning for unseen relation extraction.

On FewRel, SLM-based methods remain competitive on clustering metrics, with RoCORE tying ReaORE on ARI and ASCORE achieving the second-best V-measure F$_1$.
Among LLM-based methods, LLM-OREF obtains the highest B$^3$ F$_1$, but its classification Macro-F$_1$ is lower than ReaORE.
This indicates that a higher clustering score on a single metric does not necessarily translate into better semantic relation prediction.
ReaORE achieves the best V-measure F$_1$ and classification Macro-F$_1$, and ties the best ARI, showing its effectiveness in predicting unseen relation types.

On TACRED, SLM-based baselines show larger performance gaps, especially on classification metrics, indicating that relation extraction becomes more difficult when relation distributions are more imbalanced.
LLM-based methods improve over SLM-based methods on several metrics, but their direct prediction or clustering-oriented strategies still lag behind ReaORE.
ReaORE obtains the strongest overall performance on TACRED, achieving the best B$^3$ F$_1$, V-measure F$_1$, ARI, and classification Macro-F$_1$.
These results suggest that the coarse-to-fine reasoning process helps ReaORE maintain reliable generalization under a more challenging unseen-relation distribution.

\subsection{Ablation Study}
\begin{table}[t]
\setlength{\tabcolsep}{0.45em}
\centering
\small
\begin{tabular}{llcccc}
\toprule
\textbf{Dataset} & \textbf{Method} & \textbf{B$^3$} & \textbf{V-M} & \textbf{ARI} & \textbf{Cls.} \\
\midrule
\multirow{5}{*}{\textbf{FewRel}} 
 & w/o $\mathcal{M}_{2}$ & 79.8 & 87.4 & 73.5 & 83.2 \\
 & w/o $\mathcal{M}_{1}$+$\mathcal{M}_{2}$ & 79.8 & 84.2 & 69.0 & 81.7 \\
 & w/o $\mathcal{M}_{3}$ & 79.2 & 78.6 & 70.0 & 79.2 \\
 & w/o IR. & 78.0 & 77.7 & 67.8 & 78.0 \\
\rowcolor{gray!10} & \textbf{Ours} & \textbf{82.6} & \textbf{89.6} & \textbf{80.7} & \textbf{84.2} \\
\midrule
\multirow{5}{*}{\textbf{TACRED}} 
 & w/o $\mathcal{M}_{2}$ & 78.1 & 82.4 & 80.4 & 70.9 \\
 & w/o $\mathcal{M}_{1}$+$\mathcal{M}_{2}$ & 76.8 & 81.2 & 78.5 & 69.9 \\
 & w/o $\mathcal{M}_{3}$ & 77.0 & 81.5 & 78.8 & 69.9 \\
 & w/o IR. & 76.2 & 80.8 & 77.5 & 69.5 \\
\rowcolor{gray!10} & \textbf{Ours} & \textbf{78.8} & \textbf{83.1} & \textbf{81.3} & \textbf{71.4} \\
\bottomrule
\end{tabular}
\caption{Ablation study results on FewRel and TACRED datasets. B$^3$, V-M, and Cls. denote B$^3$ F$_1$, V-Measure F$_1$, and classification Macro-F$_1$.}
\label{tab:ablation}
\end{table}

The ablation study examines whether each part of ReaORE contributes to the intended coarse-to-fine reasoning process, with results shown in Table~\ref{tab:ablation}.
Due to space limitations, we report only F$_1$ scores for B$^3$, V-Measure, and classification metrics.
They show that ReaORE does not rely on a single strong LRM call; instead, its advantage comes from decomposing OpenRE into multi-aspect relation filtering, embedding-based relation supplementation and filtering, and fine-grained contrastive reasoning with exclusive judgment evidence.

\textbf{Effect of the relation reranking. }
To examine the effectiveness of relation reranking, we remove this step after matching-based reasoning.
In this variant, $\mathcal{M}_{1}$ still produces the score-based relation tiers, but $\mathcal{R}_c$ is constructed only from the preliminary matching results, typically using the most matched tier $\mathcal{R}_{3}$ without embedding-based supplementation.
This setting tests whether binary matching judgments alone are sufficient before fine-grained comparative reasoning.
After removing relation reranking, classification Macro-F$_1$ decreases from 84.2 to 83.2 on FewRel and from 71.4 to 70.9 on TACRED, while ARI also drops by 7.2 and 0.9 points, respectively.
These results indicate that relation reranking is useful for recovering plausible relations that may not receive the highest matching score, thereby improving the coverage of $\mathcal{R}_c$ before relation prediction.

\textbf{Effect of relation filtering.}
We further evaluate whether relation filtering is necessary before relation prediction.
The variant without both $\mathcal{M}_{1}$ and $\mathcal{M}_{2}$ removes the entire relation filtering stage and directly exposes $\mathcal{M}_{3}$ to the full unseen relation space.
Although $\mathcal{M}_{3}$ still performs fine-grained comparative reasoning, the comparison objects have not been compressed and filtered, making the relation space less concise and thus impairing performance.
As shown in Table~\ref{tab:ablation}, removing relation filtering leads to consistent drops on all reported metrics for both FewRel and TACRED.
On FewRel, B$^3$, V-Measure, ARI, and classification Macro-F$_1$ decrease to 79.8, 84.2, 69.0, and 81.7, respectively; on TACRED, they decrease to 76.8, 81.2, 78.5, and 69.9.
The performance drop indicates that fine-grained comparative reasoning still requires a compact input space: too many loosely related relations introduce noise into the comparison process, and the increased number of comparisons also lengthens the reasoning chain, which may increase the risk of relevant information being underused in long contexts~\cite{liu2024lost,he2024never}.
Therefore, relation filtering is necessary for constructing a focused $\mathcal{R}_c$ before relation prediction.

\textbf{Effect of relation prediction.}
We next examine whether relation filtering alone is sufficient for OpenRE.
The variant without $\mathcal{M}_{3}$ keeps the relation filtering stage but removes fine-grained comparative reasoning, so the final decision is made without explicit pairwise comparison among the filtered relations.
As shown in Table~\ref{tab:ablation}, removing $\mathcal{M}_{3}$ causes clear drops on both datasets.
On FewRel, V-Measure decreases from 89.6 to 78.6 and ARI decreases from 80.7 to 70.0; on TACRED, B$^3$, V-Measure, ARI, and classification Macro-F$_1$ decrease to 77.0, 81.5, 78.8, and 69.9, respectively.
These results indicate that relation filtering can narrow the relation space, but it cannot replace fine-grained comparison when the remaining relations are semantically close.
Thus, $\mathcal{M}_{3}$ is necessary for turning $\mathcal{R}_c$ into a final relation extraction decision.

\textbf{Effect of intermediate reasoning.}
We finally evaluate the role of intermediate reasoning chains.
The w/o IR. variant keeps the same overall prediction pipeline but removes the structured reasoning process, asking the model to output the final label without explicit intermediate analysis.
This setting examines whether the gains of ReaORE come only from the pipeline structure, or also from the reasoning chains generated within each stage.
As shown in Table~\ref{tab:ablation}, w/o IR. consistently underperforms the full framework on all reported metrics and both datasets.
On FewRel, classification Macro-F$_1$ decreases from 84.2 to 78.0; on TACRED, it decreases from 71.4 to 69.5.
The degradation suggests that intermediate reasoning is part of the decision process rather than a post-hoc explanation.
It makes the evidence used in relation--instance matching, relation reranking, and pairwise comparison explicit, which leads to more reliable final predictions.

\subsection{Analysis}
\begin{figure}[t]
    \centering
    \includegraphics[width=1.0\linewidth]{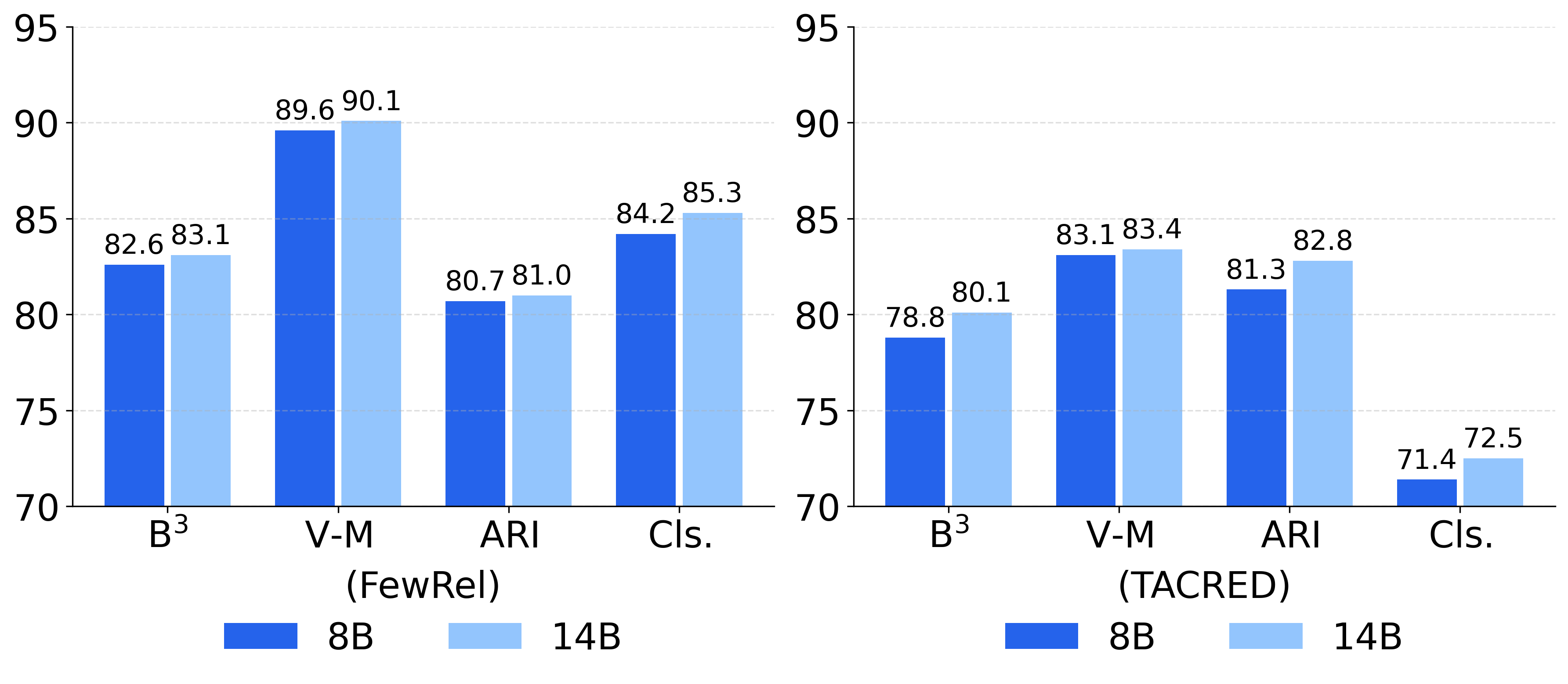}
    \caption{Performance comparison on FewRel and TACRED across different model scales.}
    \label{fig:scale_comparison}
\end{figure}

\textbf{Effectiveness on LRMs of Varying Sizes. }
We evaluate ReaORE with Qwen3-8B and Qwen3-14B backbones to study whether the framework benefits from stronger LRMs.
As shown in Figure~\ref{fig:scale_comparison}, increasing the parameter scale consistently improves performance on both FewRel and TACRED.
For classification Macro-F$_1$, the score improves from 84.2\% to 85.3\% on FewRel and from 71.4\% to 72.5\% on TACRED.
Similar gains are observed for clustering metrics, indicating that ReaORE can leverage stronger reasoning capacity across both relation discovery and relation classification views.

\textbf{Impact of Candidate Set Size. }
We further study the sensitivity of $\xi$, which denotes the number of top-ranked relations kept from $\widetilde{\mathcal{R}}_{3}$ to form $\mathcal{R}_c$ for $\mathcal{M}_{3}$.
We vary $\xi$ over $\{3,4,5,6\}$ and report the results in Table~\ref{tab:ablation_k}.

\begin{table}[t]
\centering
\setlength{\tabcolsep}{0.55em}
\small
\begin{tabular}{lccccc}
\toprule
\textbf{Dataset} & $\boldsymbol{\xi}$ & \textbf{B$^3$} & \textbf{V-M} & \textbf{ARI} & \textbf{Cls.} \\
\midrule
\multirow{4}{*}{\textbf{FewRel}} 
 & 3 & 80.5 & 88.0 & 78.0 & 81.5 \\
 & 4 & 81.8 & 88.8 & 79.5 & 83.0 \\
 & \cellcolor{gray!10}5 & \cellcolor{gray!10}\textbf{82.6} & \cellcolor{gray!10}\textbf{89.6} & \cellcolor{gray!10}\textbf{80.7} & \cellcolor{gray!10}\textbf{84.2} \\
 & 6 & 82.2 & 89.2 & 80.2 & 84.0 \\
\midrule
\multirow{4}{*}{\textbf{TACRED}} 
 & 3 & 76.0 & 80.5 & 77.5 & 69.6 \\
 & 4 & 77.5 & 81.8 & 79.5 & 70.4 \\
 & \cellcolor{gray!10}5 & \cellcolor{gray!10}\textbf{78.8} & \cellcolor{gray!10}\textbf{83.1} & \cellcolor{gray!10}\textbf{81.3} & \cellcolor{gray!10}\textbf{71.4} \\
 & 6 & 78.4 & 82.7 & 80.8 & 71.0 \\
\bottomrule
\end{tabular}
\caption{Sensitivity analysis of $\xi$, which controls the number of top-ranked relations kept in $\mathcal{R}_c$.}
\label{tab:ablation_k}
\end{table}

The results show that performance is sensitive to the size of $\mathcal{R}_c$.
When $\xi$ is too small, $\mathcal{R}_c$ may exclude the target relation before relation prediction, limiting the effectiveness of $\mathcal{M}_{3}$.
When $\xi$ is too large, more loosely related relations are introduced into pairwise comparison, increasing the difficulty of fine-grained comparative reasoning.
The best performance is achieved at $\xi=5$ on both datasets, which provides a suitable balance between coverage and compactness.

\section{Related Work}
% 传统的开放式关系抽取（OpenRE）方法主要采用基于小型语言模型（SLM）的聚类范式，将关系抽取重构为一种无监督或弱监督的表示学习与聚类问题。
Traditional OpenRE methods predominantly adopt clustering paradigms based on Small Language Models(SLMs), reformulating relation extraction as a problem of unsupervised orweakly supervisedd representation learning and clustering. 
% 例如，SelfORE~\cite{hu-etal-2020-selfore} 提出了一种自监督框架，通过自适应软聚类和迭代伪标签机制来优化上下文表示。
For instance, SelfORE~\cite{hu2020selfore} proposes a self-supervised framework that optimizes contextual representations through adaptive soft clustering and an iterative pseudo-labeling mechanism. 
% RoCORE~\cite{RoCORE_zhao2021relation} 利用已标注数据来学习面向关系的表示，从而促进对未见且未标注的关系类型的发现。
RoCORE~\cite{RoCORE_zhao2021relation} leverages labeled data to learn relation-oriented representations, facilitating the discovery of unseen, unlabeled relation types. 
% 为了进一步提升聚类质量，ASCORE~\cite{zhao-etal-2023-ascore} 引入了一种主动学习策略，动态选取具有代表性的实例进行人工标注，以此引导模型识别潜在的关系聚类。
To further improve clustering quality, ASCORE~\cite{zhao2023actively} introduces an active learning strategy that dynamically selects representative instances for manual annotation, guiding the model to identify potentially relation clusters.

% 然而，这些基于 SLM 的聚类方法仅反映了样例间的相似性，缺乏明确的语义理解。导致这些方法在处理语义相似的关系时，往往表现出较弱的判别能力和较差的泛化性能。
However, these SLM-based clustering methods merely reflect the similarities between instances and lack explicit semantic understanding. Consequently, when handling semantically similar relations, these methods often exhibit weak discriminative power and poor generalization performance.
% MixORE~\cite{mixore_wang2025towards} 将大型语言模型（LLMs）引入了开放式关系抽取（OpenRE）领域，但未能摆脱聚类范式的束缚。它依然受制于该范式的固有局限性——即单纯的聚类操作并不构成真正的关系抽取，从而使得后续的人工标注工作成为必要。
MixORE~\cite{mixore_wang2025towards} introduces LLMs into OpenRE but fails to break free from the clustering paradigm. It remains subject to the inherent limitations of this paradigm that merely performing clustering does not constitute true relation extraction, thereby necessitating manual post-annotation.
% LLM-OREF 通过提示大型语言模型（LLM），使其在给定输入实例及少量示例的情况下生成实体对之间的语义关系，从而直接预测最终的关系标签。
LLM-OREF~\cite{LLMOERF_tu2025llm} predicts final relation label directly by prompting an LLM to generate the semantic relation between entity pair given an input instance and a few demonstrations. 
% 据我们所知，本工作代表了首次尝试将大型推理模型（LRMs）引入开放域关系抽取（OpenRE）领域
To the best of our knowledge, this work represents the first attempt to introduce LRMs into the field of OpenRE.
% 我们的目标是利用 LRM 的细粒度推理能力，通过显式的多步推理链，实现对输入实例及关系标签的全面理解。
Our objective is to leverage the fine-grained reasoning capabilities of LRMs to achieve a thorough understanding of both input instances and relation labels through explicit, multi-step reasoning chains.
This enables ReaORE to distinguish between semantically similar relations, thereby effectively enhancing its generalization capabilities within complex, open-world scenarios.

\section{Conclusion}
In this paper, we proposed ReaORE, a reasoning-guided progressive framework for OpenRE that improves generalization to unseen relation types through coarse-to-fine relation reasoning.
Unlike clustering-based methods that require post-hoc relation labeling or LLM-based methods that directly generate relation labels, ReaORE decomposes OpenRE into relation filtering and relation prediction.
In relation filtering, matching-based reasoning first performs multi-aspect reasoning over sentence semantics and entity-type constraints to yield an initial relation set, and relation reranking then supplements and filters relations via embedding-based similarity to improve target-relation coverage before prediction.
The relation prediction stage performs fine-grained comparative reasoning to predict the final relation, aiming to distinguish easily confused relations through explicit pairwise judgment evidence.
To train the stage-specific reasoning abilities of the two reasoning models, we construct two reasoning-chain datasets and fine-tune them separately.
Experiments show that ReaORE outperforms existing baselines on both clustering-oriented and classification metrics, demonstrating that explicit progressive reasoning improves generalization to unseen relation types.

\section*{Limitations}
ReaORE improves unseen relation extraction through explicit progressive reasoning, but its adaptation process is still not fully dynamic.
ReaORE follows the standard OpenRE setting, where the unseen relation inventory is fixed during evaluation.
Within this setting, however, its adaptation mechanism is still relatively static: relation filtering, relation reranking, and final comparative reasoning are performed with fixed steps and a fixed filtering size, rather than being adjusted according to the difficulty of each instance.
Future work can explore more adaptive OpenRE frameworks that dynamically adjust the filtering granularity and control the depth of comparative reasoning according to instance difficulty.
Finally, the reasoning-chain training data are generated by a teacher model on known-relation instances.
Although rejection sampling is used to retain high-quality examples, the resulting reasoning behavior may still be affected by the teacher model's coverage and judgment patterns.

\bibliography{custom}

\appendix
\setcounter{figure}{0}
\setcounter{table}{0}
\renewcommand{\thefigure}{\thesection.\arabic{figure}}
\renewcommand{\thetable}{\thesection.\arabic{table}}
\makeatletter
\@addtoreset{figure}{section}
\@addtoreset{table}{section}
\makeatother

\section{Additional Experimental Details}

\subsection{Baseline Details}
\label{app:baseline_details}

We provide detailed descriptions of the baselines used in Table~\ref{tab:main}.
\textbf{SelfORE}~\cite{hu2020selfore} is a self-supervised OpenRE method that uses a pretrained language model to obtain contextualized relational features.
It performs adaptive clustering to produce pseudo labels and iteratively improves the representations through a relation classification objective.
\textbf{RoCORE}~\cite{RoCORE_zhao2021relation} is a relation-oriented clustering method that leverages labeled instances from predefined relations to learn representations more suitable for clustering relation semantics, and transfers this relational knowledge to discover unseen relations in unlabeled data.
\textbf{ASCORE}~\cite{zhao2023actively} formulates OpenRE as actively supervised clustering, where clustering learning and relation labeling are alternately performed.
It further designs an active labeling strategy to select representative instances and dynamically discover new relation clusters.

\textbf{Qwen3-8B}~\cite{yang2025qwen3} is the vanilla LLM baseline used to evaluate whether a strong backbone can directly predict unseen relations without ReaORE's progressive reasoning process.
\textbf{ORELLM}~\cite{orellm_wang2024phrases} uses cooperating LLMs for OpenRE.
It extracts relational phrases from input instances and exploits LLM-based probabilities to estimate semantic similarity between phrases, thereby supporting clustering without relying only on embedding-space distances.
\textbf{LLM-OREF}~\cite{LLMOERF_tu2025llm} directly predicts new relations with LLMs through a relation discoverer and a relation predictor.
It uses demonstrations from known relations and adopts a self-correcting inference strategy that includes relation discovery, relation denoising, and relation prediction.

\subsection{Implementation Hyperparameters}
\label{app:implementation_hyperparameters}

Table~\ref{tab:implementation_hyperparameters} summarizes the key implementation hyperparameters used in our experiments.

\begin{table}[h]
\centering
\begin{tabular}{ll}
\toprule
\textbf{Item} & \textbf{Setting} \\
\midrule
LRM backbone & Qwen3-8B \\
Embedding backbone & BGE-M3 \\
Top-$\xi$ relations & 5 \\
Training epochs & 5 \\
LoRA ($r$ / $\alpha$ / dropout) & 64 / 256 / 0.1 \\
Optimizer & AdamW \\
LR / scheduler & $1\times10^{-4}$ / cosine \\
Warmup ratio & 0.1 \\
Max sequence length & 4,096 tokens \\
Inference engine & vLLM \\
Hardware & 2 A100 80GB GPUs \\
\bottomrule
\end{tabular}
\caption{Key implementation hyperparameters used in ReaORE.}
\label{tab:implementation_hyperparameters}
\end{table}

\subsection{Comparison with MixORE}
\label{app:mixore_comparison}

We additionally compare ReaORE with MixORE~\cite{mixore_wang2025towards} under the experimental setting used in the MixORE paper.
This setting is different from the main experiments in Table~\ref{tab:main}.
MixORE emphasizes that the test data should contain a mixture of known and unseen relation types; specifically, the number of unseen relation types is set to 6, while the remaining relation types are treated as known relations.
MixORE uses BERT-base as its backbone and therefore belongs to the SLM-based OpenRE paradigm.
Table~\ref{tab:mixore_setting} reports the comparison under this setting.
The MixORE scores are taken from the MixORE paper and converted to percentage values for consistency with our tables.
For ReaORE, we report the results of the 14B backbone under the same setting.

\begin{table}[h]
\centering
\setlength{\tabcolsep}{0.5em}
\small
\begin{tabular}{llcccc}
\toprule
\textbf{Dataset} & \textbf{Method} & \textbf{B$^3$} & \textbf{V-M} & \textbf{ARI} & \textbf{Cls.} \\
\midrule
\multirow{2}{*}{FewRel}
 & MixORE & 89.7 & 88.0 & 88.2 & 83.3 \\
 & ReaORE & 90.1 & 88.8 & 88.8 & 86.0 \\
\midrule
\multirow{2}{*}{TACRED}
 & MixORE & 86.8 & 86.0 & 84.7 & 88.3 \\
 & ReaORE & 87.3 & 86.5 & 85.2 & 88.9 \\
\bottomrule
\end{tabular}
\caption{Comparison between ReaORE and MixORE under the MixORE setting. B$^3$, V-M, and Cls. denote B$^3$ F$_1$, V-Measure F$_1$, and classification Macro-F$_1$, respectively.}
\label{tab:mixore_setting}
\end{table}

\section{Visualization of Reasoning Effect}
\label{app:reasoning_visualization}

This appendix provides a qualitative error analysis to further examine how explicit reasoning affects relation prediction.
The experiment compares ReaORE with a variant that removes intermediate reasoning and directly predicts relation labels.
We focus on representative FewRel relations that are semantically close or share similar entity-type patterns, because these cases are where OpenRE models are most likely to confuse relation boundaries.

\begin{figure}[t]
    \centering
    \includegraphics[width=0.49\linewidth]{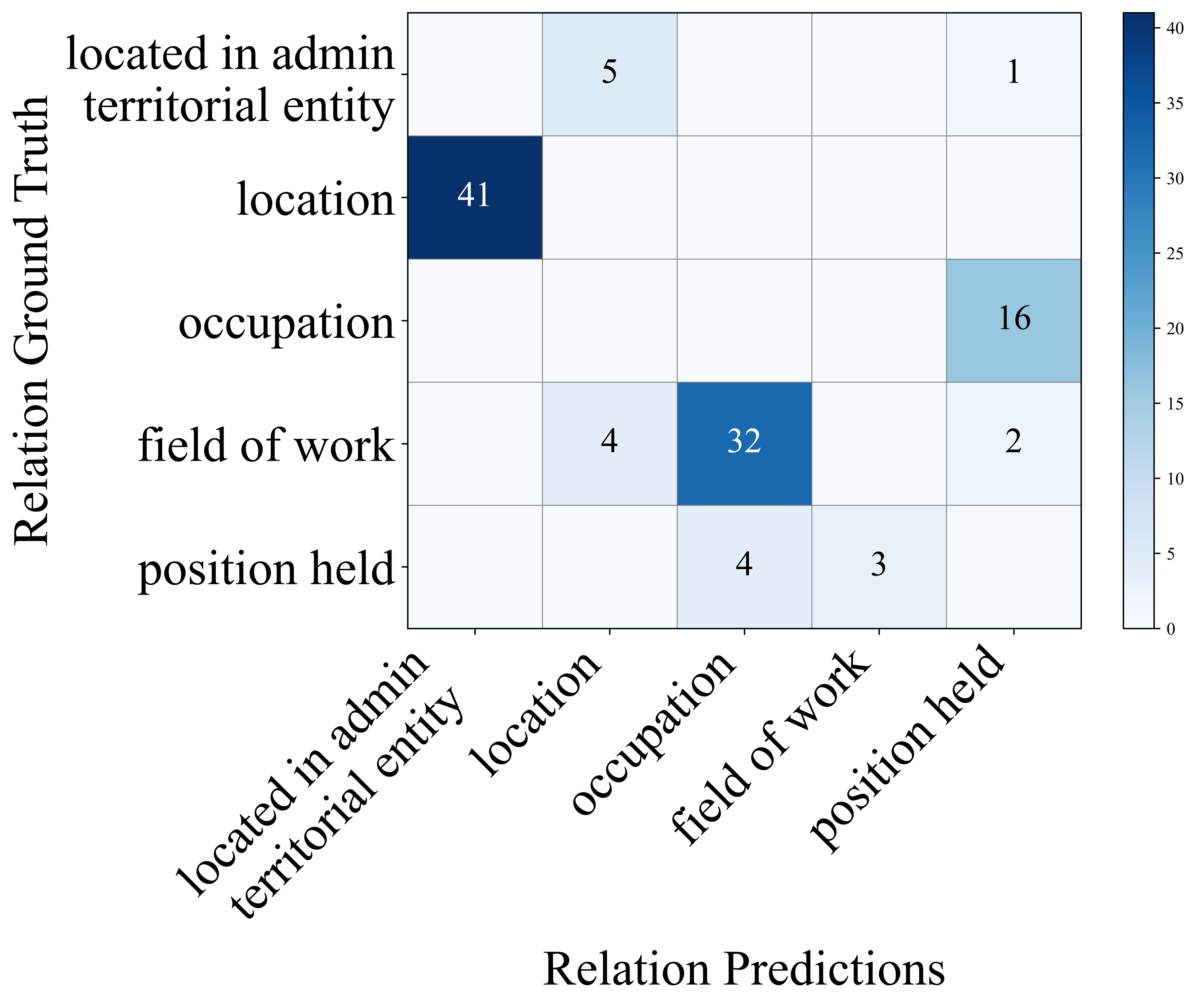}
    \hfill
    \includegraphics[width=0.49\linewidth]{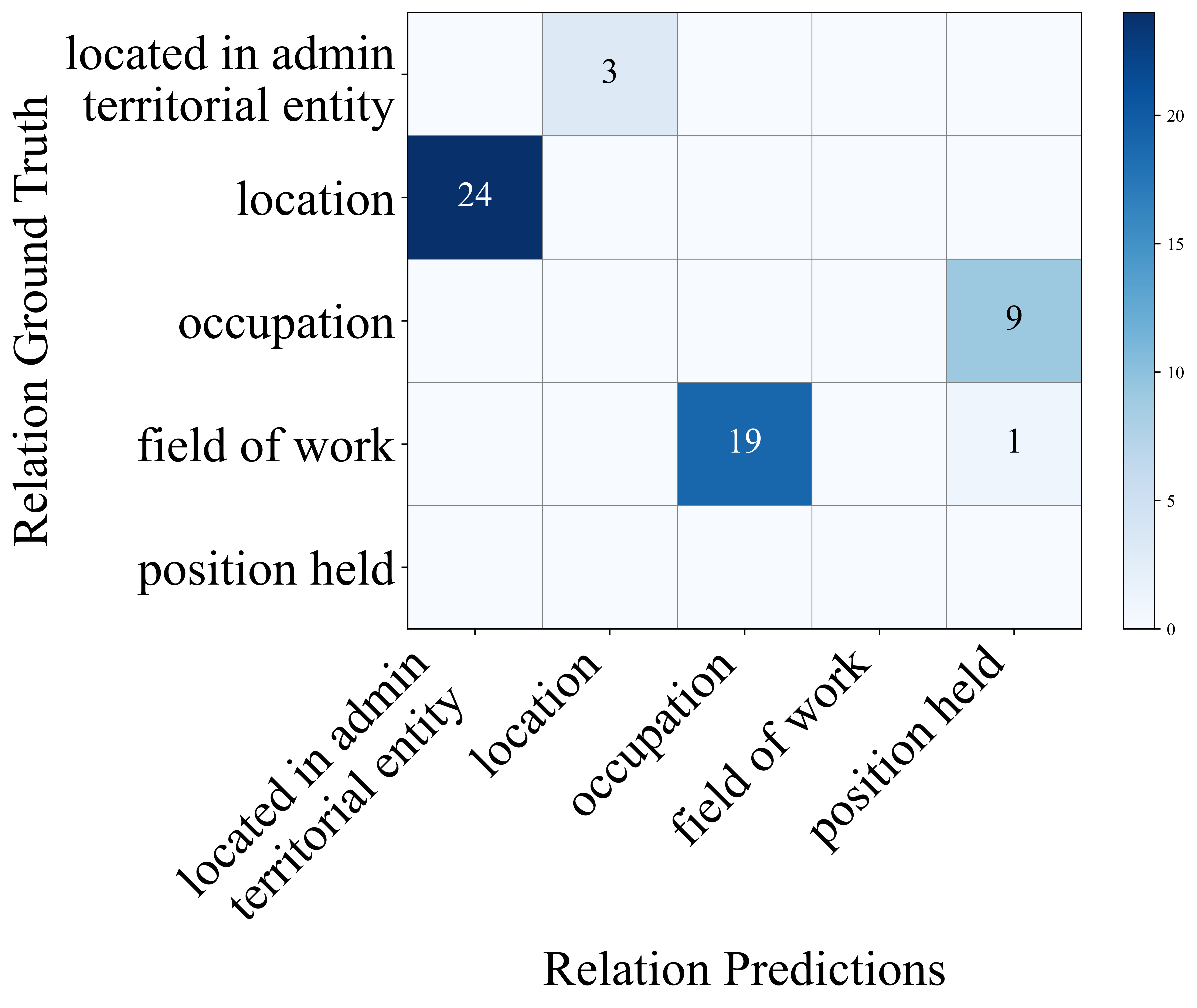}
    \caption{Error matrices on representative FewRel relations. Left: without reasoning. Right: ReaORE. Rows denote ground-truth labels and columns denote predicted labels.}
    \label{fig:error_matrix}
\end{figure}

Figure~\ref{fig:error_matrix} reports the corresponding error matrices.
Rows denote ground-truth relations and columns denote predicted relations; therefore, diagonal entries indicate correct predictions, while off-diagonal entries indicate confusions between relation types.
The left matrix shows the variant without reasoning, and the right matrix shows ReaORE.
This comparison is intended to reveal whether the matching-based reasoning and fine-grained comparative reasoning modules merely improve aggregate scores, or whether they actually reduce errors among easily confused relations.

The visualization shows that ReaORE produces fewer off-diagonal errors than the variant without reasoning.
This indicates that explicit reasoning helps the model avoid relying only on shallow lexical or entity-type similarity.
Instead, ReaORE requires the model to first check relation-instance matching from multiple aspects and then compare plausible relations before making the final decision.
As a result, the model better distinguishes subtle relation boundaries, especially for semantically adjacent relations that are difficult to separate through direct label generation.

\section{Inference Efficiency Comparison}
\label{app:inference_efficiency}

This appendix reports the inference-time cost of ReaORE.
The purpose is to clarify where the computational overhead comes from and whether explicit reasoning makes the final prediction stage inefficient.
Table~\ref{tab:inference_breakdown} summarizes the time cost of the three inference components.
Under the updated timing results, the third component is the largest single cost because $\mathcal{M}_{3}$ performs fine-grained pairwise comparisons and aggregates the comparative evidence for the final decision.
The first component remains moderate because $\mathcal{M}_{1}$ performs matching-based reasoning for each relation, but this step can be parallelized across relations.
The second component is lightweight because $\mathcal{M}_{2}$ only performs embedding-based relation reranking.

\begin{table}[t]
\centering
\setlength{\tabcolsep}{0.35em}
\small
\begin{tabular}{p{0.20\linewidth}c p{0.48\linewidth}}
\toprule
\textbf{Component} & \textbf{Time (s)} & \textbf{Description} \\
\midrule
$\mathcal{M}_{1}$ matching & 0.73 & Analyzes the query and evaluates each relation by semantic, head-entity type, and tail-entity type matching to form relation tiers. \\
$\mathcal{M}_{2}$ reranking & 0.19 & Supplements and filters relation tiers through embedding similarity, then keeps the top-ranked candidates as $\mathcal{R}_{c}$. \\
$\mathcal{M}_{3}$ prediction & 1.76 & Performs pairwise comparisons over $\mathcal{R}_{c}$ and aggregates the comparative evidence into the final relation prediction. \\
\bottomrule
\end{tabular}
\caption{Inference-time comparison across the three inference components of ReaORE.}
\label{tab:inference_breakdown}
\end{table}

We further compare different designs for the third stage in Table~\ref{tab:third_stage_efficiency}.
Removing pairwise comparison is faster, but it no longer provides explicit comparative judgments among plausible relations.
In this setting, the model uses the judgment result without two-by-two comparative reasoning; it then applies a conventional CoT-style generation process without imposing requirements on the CoT output format or the intermediate analysis procedure.
In contrast, multi-turn pairwise comparison is slower because it requires repeated interactions.
ReaORE adopts a single-dialogue comparative reasoning design, which preserves explicit pairwise judgments while keeping the additional time cost moderate.

\begin{table}[t]
\centering
\setlength{\tabcolsep}{0.35em}
\small
\begin{tabular}{p{0.28\linewidth}c p{0.43\linewidth}}
\toprule
\textbf{Design} & \textbf{Time (s)} & \textbf{Description} \\
\midrule
No pairwise comparison & 1.51 & Uses the judgment result without two-by-two comparative reasoning. \\
Pairwise comparison (ours) & 1.76 & Completes all pairwise comparisons in one dialogue. \\
Multi-turn pairwise comparison & 3.38 & Performs pairwise comparisons through multiple dialogue turns. \\
\bottomrule
\end{tabular}
\caption{Inference-time comparison across different relation prediction designs in the third stage.}
\label{tab:third_stage_efficiency}
\end{table}

\section{Prompts}
\label{app:prompts}

This appendix provides the complete prompts used by the two reasoning models in ReaORE.
The prompts define the input format, reasoning requirements, and output constraints for the two stages that require explicit reasoning.
During data construction, the same prompt formats are used to guide the teacher model in generating reasoning-chain training data.
During inference, the fine-tuned reasoning models follow the corresponding formats to produce structured intermediate reasoning.

\subsection{Matching-Based Reasoning Prompt}
We first provide the complete prompt used by the matching-based reasoning model $\mathcal{M}_{1}$ for the example shown in Figure~\ref{fig:reasoning_record_example}(a).
The purpose of this prompt is to force the model to make the evidence used for relation filtering explicit before assigning matching scores.
Specifically, the prompt asks $\mathcal{M}_{1}$ to analyze the sentence and entity pair, then evaluate each relation by semantic support, head-side argument validity, and tail-side argument validity.
The expected output is a structured reasoning trace: the \texttt{<think>} block contains the detailed query analysis and per-relation judgments, while the \texttt{<summary>} block gives the three Boolean matching scores used to form relation tiers.

\begingroup
\normalsize
\sloppy
\raggedright
\setlength{\parindent}{0pt}
\setlength{\parskip}{0.25em}
\setlength{\leftskip}{0.75em}
\noindent{}\textless{}|im\_start|\textgreater{}system\par
\noindent{}\#\#\# Role\par
\noindent{}You are an expert **Relation Extraction Specialist**. Your task is to analyze a provided text\par
\noindent{}and a pair of entities (Head and Tail) to determine which candidate relations from a given list\par
\noindent{}are correct.\par
\par
\noindent{}\#\#\# Task Logic\par
\noindent{}1.  **Analyze**: Infer entity types for **Head** and **Tail** based on Query.\par
\noindent{}2.  **Evaluate**: Check 3 criteria for each relation (1=Yes, 0=No):\par
\noindent{}    * **Semantic**: Does relation meaning match Query?\par
\noindent{}    * **Head Constraint**: Is **Head** type valid as the relation's Source/Domain? -\textgreater{} Maps to\par
\noindent{}\char96{}head\_entity\_evaluation\char96{}.\par
\noindent{}    * **Tail Constraint**: Is **Tail** type valid as the relation's Target/Range? -\textgreater{} Maps to\par
\noindent{}\char96{}tail\_entity\_evaluation\char96{}.\par
\par
\noindent{}\#\#\# Output Format\par
\noindent{}Strictly output **only** the following two parts. No markdown code blocks.\par
\par
\noindent{}**Part 1: Analysis**\par
\noindent{}Wrap in \char96{}\textless{}think\textgreater{}\char96{} tag. Use valid JSON.\par
\noindent{}\textless{}think\textgreater{}\par
\noindent{}\{\par
\noindent{}  "Query Analysis": \{\par
\noindent{}    "text\_summary": "Overview and understand this text",\par
\noindent{}    "head\_entity\_type": "Inferred type of Head Entity and explanation",\par
\noindent{}    "tail\_entity\_type": "Inferred type of Tail Entity and explanation"\par
\noindent{}  \},\par
\noindent{}  "Candidate Relation Analysis": [\par
\noindent{}    \{\par
\noindent{}      "relation\_name": "Relation Name",\par
\noindent{}      "relation\_description": "Brief analysis of relation definition",\par
\noindent{}      "semantic\_evaluation": \{\par
\noindent{}        "is\_match": 0 or 1,\par
\noindent{}        "reason": "Reason for semantic match"\par
\noindent{}      \},\par
\noindent{}      "head\_entity\_evaluation": \{\par
\noindent{}        "is\_valid": 0 or 1,\par
\noindent{}        "reason": "Is Head type valid for this relation domain?"\par
\noindent{}      \},\par
\noindent{}      "tail\_entity\_evaluation": \{\par
\noindent{}        "is\_valid": 0 or 1,\par
\noindent{}        "reason": "Is Tail type valid for this relation range?"\par
\noindent{}      \}\par
\noindent{}    \}\par
\noindent{}  ]\par
\noindent{}\}\par
\noindent{}\textless{}/think\textgreater{}\par
\par
\noindent{}**Part 2: Summary**\par
\noindent{}Wrap in \char96{}\textless{}summary\textgreater{}\char96{} tag. CSV format:\par
\noindent{}\char96{}Name,Semantic(1/0),\char96{}\par
\noindent{}\char96{}HeadValid(1/0),TailValid(1/0)\char96{}.\par
\noindent{}\textless{}summary\textgreater{}\par
\noindent{}RelationName1,1,1,0\par
\noindent{}RelationName2,0,1,1\par
\noindent{}\textless{}/summary\textgreater{}\par
\noindent{}\textless{}|im\_end|\textgreater{}\par
\noindent{}\textless{}|im\_start|\textgreater{}user\par
\noindent{}Text: It is an affiliated member of the Local Government Association of England and Wales and\par
\noindent{}the regional employers organisation .\par
\noindent{}Head entity: local government association.\par
\noindent{}Tail entity: england and wales.\par
\endgroup

\subsection{Fine-Grained Comparative Reasoning Prompt}
We next provide the complete prompt used by the contrastive reasoning model $\mathcal{M}_{3}$ for the example shown in Figure~\ref{fig:reasoning_record_example}(b).
After relation filtering, the remaining relations are plausible but can still be semantically close.
The goal of this prompt is therefore not to score each relation independently, but to require pairwise comparison under explicit decision rules.
The prompt asks $\mathcal{M}_{3}$ to prioritize sentence evidence, prefer the more precise relation when multiple relations are plausible, and align the final answer with the phrase or event expressed in the sentence.
The expected output contains a \texttt{comparison\_log} for pairwise decisions and a \texttt{final\_result} that states the selected relation and decision logic.

\begingroup
\normalsize
\sloppy
\raggedright
\setlength{\parindent}{0pt}
\setlength{\parskip}{0.25em}
\setlength{\leftskip}{0.75em}
\noindent{}\textless{}|im\_start|\textgreater{}system\par
\noindent{}You are a relation extraction expert. Select the best relation from candidate relations.\par
\par
\noindent{}Rules:\par
\noindent{}1. Evidence priority: Does the text clearly support this relation?\par
\noindent{}2. Precision: If both are correct, choose the more specific one\par
\noindent{}3. Context matching: Which relation is closest to the verb/phrase in the text?\par
\par
\noindent{}Output format:\par
\noindent{}\textless{}think\textgreater{}\par
\noindent{}\{\par
\noindent{}  "comparison\_log": [\par
\noindent{}    \{"pair": "A vs B", "winner": "A", "reason": "reason"\},\par
\noindent{}    ...\par
\noindent{}  ],\par
\noindent{}  "final\_result": \{\par
\noindent{}    "best\_relation": "best relation name",\par
\noindent{}    "decision\_logic": "selection logic"\par
\noindent{}  \}\par
\noindent{}\}\par
\noindent{}\textless{}/think\textgreater{}\par
\par
\noindent{}\textless{}summary\textgreater{}\par
\noindent{}best relation name\par
\noindent{}\textless{}/summary\textgreater{}\textless{}|im\_end|\textgreater{}\par
\noindent{}\textless{}|im\_start|\textgreater{}user\par
\noindent{}Text: It is an affiliated member of the Local Government Association of England and Wales and\par
\noindent{}the regional employers organisation .\par
\noindent{}Head entity: local government association.\par
\noindent{}Tail entity: england and wales.\par
\par
\noindent{}**Candidate Relations to Compare:**\par
\noindent{}- applies to jurisdiction\par
\noindent{}- located in the administrative territorial entity\par
\noindent{}- location\par
\noindent{}- contains administrative territorial entity\par
\par
\noindent{}Please compare these relations pairwise and identify the single best relation.\textless{}|im\_end|\textgreater{}\par
\noindent{}\textless{}|im\_start|\textgreater{}assistant\par
\endgroup

\section{Reasoning Trace Examples}
\label{app:reasoning_traces}

This appendix provides the complete reasoning traces generated from the prompts in Appendix~\ref{app:prompts}.
While Figure~\ref{fig:reasoning_record_example} in the main paper only shows abbreviated examples, the material here presents the full intermediate reasoning for one representative FewRel instance.
The goal is to make the reasoning procedure inspectable: readers can see how the models analyze evidence, assign intermediate judgments, and produce the final relation decision.

\subsection{Matching-Based Reasoning Trace}
Given the matching-based reasoning prompt, $\mathcal{M}_{1}$ produces the following reasoning trace.
This trace illustrates how ReaORE converts an input sentence and entity pair into explicit multi-aspect judgments over relations.
The reader should focus on two parts: the query analysis, which identifies the sentence evidence and entity types, and the relation analysis, which explains why each relation receives its semantic, head-entity, and tail-entity matching scores.
These scores are the basis for constructing the initial relation tiers used by the subsequent reranking step.

\begingroup
\normalsize
\sloppy
\raggedright
\setlength{\parindent}{0pt}
\setlength{\parskip}{0pt}
\setlength{\leftskip}{0.75em}
\noindent{}----------------------------------------\par
\noindent{}Query Index: 0\par
\noindent{}----------------------------------------\par
\par
\noindent{}Input Text:\par
\noindent{}Text: It is an affiliated member of the Local Government Association of England and Wales and the regional employers organisation .\par
\noindent{}Head entity: local government association.\par
\noindent{}Tail entity: england and wales.\par
\par
\noindent{}----------------------------------------\par
\noindent{}Raw Output:\par
\noindent{}----------------------------------------\par
\par
\noindent{}<think>\par
\noindent{}\{\par
\noindent{}\hspace*{0.70em}"Query Analysis": \{\par
\noindent{}\hspace*{1.40em}"text\_summary": "The text states that an unspecified entity is a member of the 'Local Government Association of England and Wales'. This implies a relationship between the association and the geographical region it serves, which is England and Wales.",\par
\noindent{}\hspace*{1.40em}"head\_entity\_type": "The head entity 'local government association' is an organization, specifically a professional or trade association.",\par
\noindent{}\hspace*{1.40em}"tail\_entity\_type": "The tail entity 'england and wales' is a geographical and political region, a country."\par
\noindent{}\hspace*{0.70em}\},\par
\noindent{}\hspace*{0.70em}"Candidate Relation Analysis": [\par
\noindent{}\hspace*{1.40em}\{\par
\noindent{}\hspace*{2.10em}"relation\_name": "residence",\par
\noindent{}\hspace*{2.10em}"relation\_description": "The place where a person is or has been resident.",\par
\noindent{}\hspace*{2.10em}"semantic\_evaluation": \{\par
\noindent{}\hspace*{2.80em}"is\_match": 0,\par
\noindent{}\hspace*{2.80em}"reason": "The text describes an organizational relationship, not a person's place of residence."\par
\noindent{}\hspace*{2.10em}\},\par
\noindent{}\hspace*{2.10em}"head\_entity\_evaluation": \{\par
\noindent{}\hspace*{2.80em}"is\_valid": 0,\par
\noindent{}\hspace*{2.80em}"reason": "The head entity is an organization, but the relation's domain is a person."\par
\noindent{}\hspace*{2.10em}\},\par
\noindent{}\hspace*{2.10em}"tail\_entity\_evaluation": \{\par
\noindent{}\hspace*{2.80em}"is\_valid": 1,\par
\noindent{}\hspace*{2.80em}"reason": "The tail entity is a country/place, which is a valid type for this relation's range."\par
\noindent{}\hspace*{2.10em}\}\par
\noindent{}\hspace*{1.40em}\},\par
\noindent{}\hspace*{1.40em}\{\par
\noindent{}\hspace*{2.10em}"relation\_name": "headquarters location",\par
\noindent{}\hspace*{2.10em}"relation\_description": "The specific location where an organization's headquarters is or has been situated.",\par
\noindent{}\hspace*{2.10em}"semantic\_evaluation": \{\par
\noindent{}\hspace*{2.80em}"is\_match": 0,\par
\noindent{}\hspace*{2.80em}"reason": "The text does not state that the headquarters of the Local Government Association is in England and Wales. It implies the association serves that region."\par
\noindent{}\hspace*{2.10em}\},\par
\noindent{}\hspace*{2.10em}"head\_entity\_evaluation": \{\par
\noindent{}\hspace*{2.80em}"is\_valid": 1,\par
\noindent{}\hspace*{2.80em}"reason": "The head entity is an organization, which is a valid type for this relation's domain."\par
\noindent{}\hspace*{2.10em}\},\par
\noindent{}\hspace*{2.10em}"tail\_entity\_evaluation": \{\par
\noindent{}\hspace*{2.80em}"is\_valid": 1,\par
\noindent{}\hspace*{2.80em}"reason": "The tail entity is a country, which can be a valid location for a headquarters."\par
\noindent{}\hspace*{2.10em}\}\par
\noindent{}\hspace*{1.40em}\},\par
\noindent{}\hspace*{1.40em}\{\par
\noindent{}\hspace*{2.10em}"relation\_name": "military rank",\par
\noindent{}\hspace*{2.10em}"relation\_description": "A military rank achieved by a person.",\par
\noindent{}\hspace*{2.10em}"semantic\_evaluation": \{\par
\noindent{}\hspace*{2.80em}"is\_match": 0,\par
\noindent{}\hspace*{2.80em}"reason": "The text is about civil government and has no connection to the military."\par
\noindent{}\hspace*{2.10em}\},\par
\noindent{}\hspace*{2.10em}"head\_entity\_evaluation": \{\par
\noindent{}\hspace*{2.80em}"is\_valid": 0,\par
\noindent{}\hspace*{2.80em}"reason": "The head entity is a civil organization, not a person."\par
\noindent{}\hspace*{2.10em}\},\par
\noindent{}\hspace*{2.10em}"tail\_entity\_evaluation": \{\par
\noindent{}\hspace*{2.80em}"is\_valid": 0,\par
\noindent{}\hspace*{2.80em}"reason": "The tail entity is a country, not a military rank."\par
\noindent{}\hspace*{2.10em}\}\par
\noindent{}\hspace*{1.40em}\},\par
\noindent{}\hspace*{1.40em}\{\par
\noindent{}\hspace*{2.10em}"relation\_name": "child",\par
\noindent{}\hspace*{2.10em}"relation\_description": "The subject has the object as their biological, foster, or adoptive child.",\par
\noindent{}\hspace*{2.10em}"semantic\_evaluation": \{\par
\noindent{}\hspace*{2.80em}"is\_match": 0,\par
\noindent{}\hspace*{2.80em}"reason": "The entities are an organization and a country, not people with a familial relationship."\par
\noindent{}\hspace*{2.10em}\},\par
\noindent{}\hspace*{2.10em}"head\_entity\_evaluation": \{\par
\noindent{}\hspace*{2.80em}"is\_valid": 0,\par
\noindent{}\hspace*{2.80em}"reason": "The head entity is an organization, not a person."\par
\noindent{}\hspace*{2.10em}\},\par
\noindent{}\hspace*{2.10em}"tail\_entity\_evaluation": \{\par
\noindent{}\hspace*{2.80em}"is\_valid": 0,\par
\noindent{}\hspace*{2.80em}"reason": "The tail entity is a country, not a person."\par
\noindent{}\hspace*{2.10em}\}\par
\noindent{}\hspace*{1.40em}\},\par
\noindent{}\hspace*{1.40em}\{\par
\noindent{}\hspace*{2.10em}"relation\_name": "operator",\par
\noindent{}\hspace*{2.10em}"relation\_description": "The person, profession, or organization that operates the equipment, facility, or service.",\par
\noindent{}\hspace*{2.10em}"semantic\_evaluation": \{\par
\noindent{}\hspace*{2.80em}"is\_match": 0,\par
\noindent{}\hspace*{2.80em}"reason": "The text does not state that the Local Government Association operates England and Wales."\par
\noindent{}\hspace*{2.10em}\},\par
\noindent{}\hspace*{2.10em}"head\_entity\_evaluation": \{\par
\noindent{}\hspace*{2.80em}"is\_valid": 1,\par
\noindent{}\hspace*{2.80em}"reason": "The head entity is an organization, which could be an operator."\par
\noindent{}\hspace*{2.10em}\},\par
\noindent{}\hspace*{2.10em}"tail\_entity\_evaluation": \{\par
\noindent{}\hspace*{2.80em}"is\_valid": 0,\par
\noindent{}\hspace*{2.80em}"reason": "The tail entity is a country, not an equipment, facility, or service."\par
\noindent{}\hspace*{2.10em}\}\par
\noindent{}\hspace*{1.40em}\},\par
\noindent{}\hspace*{1.40em}\{\par
\noindent{}\hspace*{2.10em}"relation\_name": "heritage designation",\par
\noindent{}\hspace*{2.10em}"relation\_description": "The heritage designation of a cultural or natural site.",\par
\noindent{}\hspace*{2.10em}"semantic\_evaluation": \{\par
\noindent{}\hspace*{2.80em}"is\_match": 0,\par
\noindent{}\hspace*{2.80em}"reason": "The text is about an association, not a heritage site."\par
\noindent{}\hspace*{2.10em}\},\par
\noindent{}\hspace*{2.10em}"head\_entity\_evaluation": \{\par
\noindent{}\hspace*{2.80em}"is\_valid": 0,\par
\noindent{}\hspace*{2.80em}"reason": "The head entity is an organization, not a cultural or natural site."\par
\noindent{}\hspace*{2.10em}\},\par
\noindent{}\hspace*{2.10em}"tail\_entity\_evaluation": \{\par
\noindent{}\hspace*{2.80em}"is\_valid": 0,\par
\noindent{}\hspace*{2.80em}"reason": "The tail entity is a country, not a heritage designation."\par
\noindent{}\hspace*{2.10em}\}\par
\noindent{}\hspace*{1.40em}\},\par
\noindent{}\hspace*{1.40em}\{\par
\noindent{}\hspace*{2.10em}"relation\_name": "field of work",\par
\noindent{}\hspace*{2.10em}"relation\_description": "The specialization of a person or organization.",\par
\noindent{}\hspace*{2.10em}"semantic\_evaluation": \{\par
\noindent{}\hspace*{2.80em}"is\_match": 0,\par
\noindent{}\hspace*{2.80em}"reason": "The text does not state that England and Wales is the field of work for the Local Government Association."\par
\noindent{}\hspace*{2.10em}\},\par
\noindent{}\hspace*{2.10em}"head\_entity\_evaluation": \{\par
\noindent{}\hspace*{2.80em}"is\_valid": 1,\par
\noindent{}\hspace*{2.80em}"reason": "The head entity is an organization, which can have a field of work."\par
\noindent{}\hspace*{2.10em}\},\par
\noindent{}\hspace*{2.10em}"tail\_entity\_evaluation": \{\par
\noindent{}\hspace*{2.80em}"is\_valid": 0,\par
\noindent{}\hspace*{2.80em}"reason": "The tail entity is a country, not a field of work."\par
\noindent{}\hspace*{2.10em}\}\par
\noindent{}\hspace*{1.40em}\},\par
\noindent{}\hspace*{1.40em}\{\par
\noindent{}\hspace*{2.10em}"relation\_name": "position played on team / speciality",\par
\noindent{}\hspace*{2.10em}"relation\_description": "The position or specialism of a player on a team.",\par
\noindent{}\hspace*{2.10em}"semantic\_evaluation": \{\par
\noindent{}\hspace*{2.80em}"is\_match": 0,\par
\noindent{}\hspace*{2.80em}"reason": "The entities are not a player and a team."\par
\noindent{}\hspace*{2.10em}\},\par
\noindent{}\hspace*{2.10em}"head\_entity\_evaluation": \{\par
\noindent{}\hspace*{2.80em}"is\_valid": 0,\par
\noindent{}\hspace*{2.80em}"reason": "The head entity is an organization, not a player."\par
\noindent{}\hspace*{2.10em}\},\par
\noindent{}\hspace*{2.10em}"tail\_entity\_evaluation": \{\par
\noindent{}\hspace*{2.80em}"is\_valid": 0,\par
\noindent{}\hspace*{2.80em}"reason": "The tail entity is a country, not a team or position."\par
\noindent{}\hspace*{2.10em}\}\par
\noindent{}\hspace*{1.40em}\},\par
\noindent{}\hspace*{1.40em}\{\par
\noindent{}\hspace*{2.10em}"relation\_name": "characters",\par
\noindent{}\hspace*{2.10em}"relation\_description": "The characters which appear in this item (like plays, operas, operettas, books, comics, films, TV series, video games).",\par
\noindent{}\hspace*{2.10em}"semantic\_evaluation": \{\par
\noindent{}\hspace*{2.80em}"is\_match": 0,\par
\noindent{}\hspace*{2.80em}"reason": "The entities are not a creative work and characters."\par
\noindent{}\hspace*{2.10em}\},\par
\noindent{}\hspace*{2.10em}"head\_entity\_evaluation": \{\par
\noindent{}\hspace*{2.80em}"is\_valid": 0,\par
\noindent{}\hspace*{2.80em}"reason": "The head entity is an organization, not a creative work."\par
\noindent{}\hspace*{2.10em}\},\par
\noindent{}\hspace*{2.10em}"tail\_entity\_evaluation": \{\par
\noindent{}\hspace*{2.80em}"is\_valid": 0,\par
\noindent{}\hspace*{2.80em}"reason": "The tail entity is a country, not a character."\par
\noindent{}\hspace*{2.10em}\}\par
\noindent{}\hspace*{1.40em}\},\par
\noindent{}\hspace*{1.40em}\{\par
\noindent{}\hspace*{2.10em}"relation\_name": "applies to jurisdiction",\par
\noindent{}\hspace*{2.10em}"relation\_description": "The item (an institution, law, public office ...) or statement belongs to or has power over or applies to the value (a territorial jurisdiction: a country, state, municipality, ...).",\par
\noindent{}\hspace*{2.10em}"semantic\_evaluation": \{\par
\noindent{}\hspace*{2.80em}"is\_match": 1,\par
\noindent{}\hspace*{2.80em}"reason": "The text implies that the Local Government Association (an institution) operates within the jurisdiction of England and Wales. This relation captures that an organization applies to or has power over a jurisdiction."\par
\noindent{}\hspace*{2.10em}\},\par
\noindent{}\hspace*{2.10em}"head\_entity\_evaluation": \{\par
\noindent{}\hspace*{2.80em}"is\_valid": 1,\par
\noindent{}\hspace*{2.80em}"reason": "The head entity is an organization, which fits the description of an 'institution' or 'public office'."\par
\noindent{}\hspace*{2.10em}\},\par
\noindent{}\hspace*{2.10em}"tail\_entity\_evaluation": \{\par
\noindent{}\hspace*{2.80em}"is\_valid": 1,\par
\noindent{}\hspace*{2.80em}"reason": "The tail entity is a country, which is a 'territorial jurisdiction'."\par
\noindent{}\hspace*{2.10em}\}\par
\noindent{}\hspace*{1.40em}\},\par
\noindent{}\hspace*{1.40em}\{\par
\noindent{}\hspace*{2.10em}"relation\_name": "sibling",\par
\noindent{}\hspace*{2.10em}"relation\_description": "The subject has the object as their sibling.",\par
\noindent{}\hspace*{2.10em}"semantic\_evaluation": \{\par
\noindent{}\hspace*{2.80em}"is\_match": 0,\par
\noindent{}\hspace*{2.80em}"reason": "The entities are an organization and a country, not people with a sibling relationship."\par
\noindent{}\hspace*{2.10em}\},\par
\noindent{}\hspace*{2.10em}"head\_entity\_evaluation": \{\par
\noindent{}\hspace*{2.80em}"is\_valid": 0,\par
\noindent{}\hspace*{2.80em}"reason": "The head entity is an organization, not a person."\par
\noindent{}\hspace*{2.10em}\},\par
\noindent{}\hspace*{2.10em}"tail\_entity\_evaluation": \{\par
\noindent{}\hspace*{2.80em}"is\_valid": 0,\par
\noindent{}\hspace*{2.80em}"reason": "The tail entity is a country, not a person."\par
\noindent{}\hspace*{2.10em}\}\par
\noindent{}\hspace*{1.40em}\},\par
\noindent{}\hspace*{1.40em}\{\par
\noindent{}\hspace*{2.10em}"relation\_name": "sport",\par
\noindent{}\hspace*{2.10em}"relation\_description": "The sport in which the subject participates or belongs to.",\par
\noindent{}\hspace*{2.10em}"semantic\_evaluation": \{\par
\noindent{}\hspace*{2.80em}"is\_match": 0,\par
\noindent{}\hspace*{2.80em}"reason": "The text is not about sports."\par
\noindent{}\hspace*{2.10em}\},\par
\noindent{}\hspace*{2.10em}"head\_entity\_evaluation": \{\par
\noindent{}\hspace*{2.80em}"is\_valid": 0,\par
\noindent{}\hspace*{2.80em}"reason": "The head entity is an organization, not a person or team."\par
\noindent{}\hspace*{2.10em}\},\par
\noindent{}\hspace*{2.10em}"tail\_entity\_evaluation": \{\par
\noindent{}\hspace*{2.80em}"is\_valid": 0,\par
\noindent{}\hspace*{2.80em}"reason": "The tail entity is a country, not a sport."\par
\noindent{}\hspace*{2.10em}\}\par
\noindent{}\hspace*{1.40em}\},\par
\noindent{}\hspace*{1.40em}\{\par
\noindent{}\hspace*{2.10em}"relation\_name": "member of political party",\par
\noindent{}\hspace*{2.10em}"relation\_description": "The political party of which this politician is or has been a member.",\par
\noindent{}\hspace*{2.10em}"semantic\_evaluation": \{\par
\noindent{}\hspace*{2.80em}"is\_match": 0,\par
\noindent{}\hspace*{2.80em}"reason": "The text is about an association, not a political party."\par
\noindent{}\hspace*{2.10em}\},\par
\noindent{}\hspace*{2.10em}"head\_entity\_evaluation": \{\par
\noindent{}\hspace*{2.80em}"is\_valid": 0,\par
\noindent{}\hspace*{2.80em}"reason": "The head entity is an organization, not a politician."\par
\noindent{}\hspace*{2.10em}\},\par
\noindent{}\hspace*{2.10em}"tail\_entity\_evaluation": \{\par
\noindent{}\hspace*{2.80em}"is\_valid": 0,\par
\noindent{}\hspace*{2.80em}"reason": "The tail entity is a country, not a political party."\par
\noindent{}\hspace*{2.10em}\}\par
\noindent{}\hspace*{1.40em}\},\par
\noindent{}\hspace*{1.40em}\{\par
\noindent{}\hspace*{2.10em}"relation\_name": "position held",\par
\noindent{}\hspace*{2.10em}"relation\_description": "The subject currently or formerly holds the object position or public office.",\par
\noindent{}\hspace*{2.10em}"semantic\_evaluation": \{\par
\noindent{}\hspace*{2.80em}"is\_match": 0,\par
\noindent{}\hspace*{2.80em}"reason": "The text does not state that the Local Government Association holds a position called 'England and Wales'."\par
\noindent{}\hspace*{2.10em}\},\par
\noindent{}\hspace*{2.10em}"head\_entity\_evaluation": \{\par
\noindent{}\hspace*{2.80em}"is\_valid": 1,\par
\noindent{}\hspace*{2.80em}"reason": "The head entity is an organization, which can hold a position."\par
\noindent{}\hspace*{2.10em}\},\par
\noindent{}\hspace*{2.10em}"tail\_entity\_evaluation": \{\par
\noindent{}\hspace*{2.80em}"is\_valid": 0,\par
\noindent{}\hspace*{2.80em}"reason": "The tail entity is a country, not a position or public office."\par
\noindent{}\hspace*{2.10em}\}\par
\noindent{}\hspace*{1.40em}\},\par
\noindent{}\hspace*{1.40em}\{\par
\noindent{}\hspace*{2.10em}"relation\_name": "notable work",\par
\noindent{}\hspace*{2.10em}"relation\_description": "A notable scientific, artistic or literary work, or other work of significance among the subject's works.",\par
\noindent{}\hspace*{2.10em}"semantic\_evaluation": \{\par
\noindent{}\hspace*{2.80em}"is\_match": 0,\par
\noindent{}\hspace*{2.80em}"reason": "The text does not describe England and Wales as a work of the Local Government Association."\par
\noindent{}\hspace*{2.10em}\},\par
\noindent{}\hspace*{2.10em}"head\_entity\_evaluation": \{\par
\noindent{}\hspace*{2.80em}"is\_valid": 1,\par
\noindent{}\hspace*{2.80em}"reason": "The head entity is an organization, which can have notable works."\par
\noindent{}\hspace*{2.10em}\},\par
\noindent{}\hspace*{2.10em}"tail\_entity\_evaluation": \{\par
\noindent{}\hspace*{2.80em}"is\_valid": 0,\par
\noindent{}\hspace*{2.80em}"reason": "The tail entity is a country, not a work."\par
\noindent{}\hspace*{2.10em}\}\par
\noindent{}\hspace*{1.40em}\},\par
\noindent{}\hspace*{1.40em}\{\par
\noindent{}\hspace*{2.10em}"relation\_name": "voice type",\par
\noindent{}\hspace*{2.10em}"relation\_description": "A person's voice type.",\par
\noindent{}\hspace*{2.10em}"semantic\_evaluation": \{\par
\noindent{}\hspace*{2.80em}"is\_match": 0,\par
\noindent{}\hspace*{2.80em}"reason": "The text is not about music or voice types."\par
\noindent{}\hspace*{2.10em}\},\par
\noindent{}\hspace*{2.10em}"head\_entity\_evaluation": \{\par
\noindent{}\hspace*{2.80em}"is\_valid": 0,\par
\noindent{}\hspace*{2.80em}"reason": "The head entity is an organization, not a person."\par
\noindent{}\hspace*{2.10em}\},\par
\noindent{}\hspace*{2.10em}"tail\_entity\_evaluation": \{\par
\noindent{}\hspace*{2.80em}"is\_valid": 0,\par
\noindent{}\hspace*{2.80em}"reason": "The tail entity is a country, not a voice type."\par
\noindent{}\hspace*{2.10em}\}\par
\noindent{}\hspace*{1.40em}\},\par
\noindent{}\hspace*{1.40em}\{\par
\noindent{}\hspace*{2.10em}"relation\_name": "screenwriter",\par
\noindent{}\hspace*{2.10em}"relation\_description": "The person(s) who wrote the script for the subject item.",\par
\noindent{}\hspace*{2.10em}"semantic\_evaluation": \{\par
\noindent{}\hspace*{2.80em}"is\_match": 0,\par
\noindent{}\hspace*{2.80em}"reason": "The text is not about films or screenwriting."\par
\noindent{}\hspace*{2.10em}\},\par
\noindent{}\hspace*{2.10em}"head\_entity\_evaluation": \{\par
\noindent{}\hspace*{2.80em}"is\_valid": 0,\par
\noindent{}\hspace*{2.80em}"reason": "The head entity is an organization, not a creative work."\par
\noindent{}\hspace*{2.10em}\},\par
\noindent{}\hspace*{2.10em}"tail\_entity\_evaluation": \{\par
\noindent{}\hspace*{2.80em}"is\_valid": 0,\par
\noindent{}\hspace*{2.80em}"reason": "The tail entity is a country, not a person."\par
\noindent{}\hspace*{2.10em}\}\par
\noindent{}\hspace*{1.40em}\},\par
\noindent{}\hspace*{1.40em}\{\par
\noindent{}\hspace*{2.10em}"relation\_name": "contains administrative territorial entity",\par
\noindent{}\hspace*{2.10em}"relation\_description": "The list of direct subdivisions of an administrative territorial entity.",\par
\noindent{}\hspace*{2.10em}"semantic\_evaluation": \{\par
\noindent{}\hspace*{2.80em}"is\_match": 0,\par
\noindent{}\hspace*{2.80em}"reason": "The text does not state that the Local Government Association contains England and Wales as a subdivision."\par
\noindent{}\hspace*{2.10em}\},\par
\noindent{}\hspace*{2.10em}"head\_entity\_evaluation": \{\par
\noindent{}\hspace*{2.80em}"is\_valid": 0,\par
\noindent{}\hspace*{2.80em}"reason": "The head entity is an organization, not an administrative territorial entity."\par
\noindent{}\hspace*{2.10em}\},\par
\noindent{}\hspace*{2.10em}"tail\_entity\_evaluation": \{\par
\noindent{}\hspace*{2.80em}"is\_valid": 1,\par
\noindent{}\hspace*{2.80em}"reason": "The tail entity is a country, which is an administrative territorial entity."\par
\noindent{}\hspace*{2.10em}\}\par
\noindent{}\hspace*{1.40em}\},\par
\noindent{}\hspace*{1.40em}\{\par
\noindent{}\hspace*{2.10em}"relation\_name": "location of formation",\par
\noindent{}\hspace*{2.10em}"relation\_description": "The location where a group or organization was formed.",\par
\noindent{}\hspace*{2.10em}"semantic\_evaluation": \{\par
\noindent{}\hspace*{2.80em}"is\_match": 0,\par
\noindent{}\hspace*{2.80em}"reason": "The text does not state that the Local Government Association was formed in England and Wales."\par
\noindent{}\hspace*{2.10em}\},\par
\noindent{}\hspace*{2.10em}"head\_entity\_evaluation": \{\par
\noindent{}\hspace*{2.80em}"is\_valid": 1,\par
\noindent{}\hspace*{2.80em}"reason": "The head entity is an organization, which can have a location of formation."\par
\noindent{}\hspace*{2.10em}\},\par
\noindent{}\hspace*{2.10em}"tail\_entity\_evaluation": \{\par
\noindent{}\hspace*{2.80em}"is\_valid": 1,\par
\noindent{}\hspace*{2.80em}"reason": "The tail entity is a country, which is a valid location."\par
\noindent{}\hspace*{2.10em}\}\par
\noindent{}\hspace*{1.40em}\},\par
\noindent{}\hspace*{1.40em}\{\par
\noindent{}\hspace*{2.10em}"relation\_name": "located in the administrative territorial entity",\par
\noindent{}\hspace*{2.10em}"relation\_description": "The item is located on the territory of the following administrative entity.",\par
\noindent{}\hspace*{2.10em}"semantic\_evaluation": \{\par
\noindent{}\hspace*{2.80em}"is\_match": 0,\par
\noindent{}\hspace*{2.80em}"reason": "The text describes the association's area of operation, not its physical location. The relation is about location, not jurisdiction or service area."\par
\noindent{}\hspace*{2.10em}\},\par
\noindent{}\hspace*{2.10em}"head\_entity\_evaluation": \{\par
\noindent{}\hspace*{2.80em}"is\_valid": 1,\par
\noindent{}\hspace*{2.80em}"reason": "The head entity is an organization, which can be located in a place."\par
\noindent{}\hspace*{2.10em}\},\par
\noindent{}\hspace*{2.10em}"tail\_entity\_evaluation": \{\par
\noindent{}\hspace*{2.80em}"is\_valid": 1,\par
\noindent{}\hspace*{2.80em}"reason": "The tail entity is a country, which is an administrative territorial entity."\par
\noindent{}\hspace*{2.10em}\}\par
\noindent{}\hspace*{1.40em}\},\par
\noindent{}\hspace*{1.40em}\{\par
\noindent{}\hspace*{2.10em}"relation\_name": "said to be the same as",\par
\noindent{}\hspace*{2.10em}"relation\_description": "This item is said to be the same as that item, but the statement is disputed.",\par
\noindent{}\hspace*{2.10em}"semantic\_evaluation": \{\par
\noindent{}\hspace*{2.80em}"is\_match": 0,\par
\noindent{}\hspace*{2.80em}"reason": "The Local Government Association and England and Wales are distinct entities."\par
\noindent{}\hspace*{2.10em}\},\par
\noindent{}\hspace*{2.10em}"head\_entity\_evaluation": \{\par
\noindent{}\hspace*{2.80em}"is\_valid": 1,\par
\noindent{}\hspace*{2.80em}"reason": "The head entity is an organization, which can be an item."\par
\noindent{}\hspace*{2.10em}\},\par
\noindent{}\hspace*{2.10em}"tail\_entity\_evaluation": \{\par
\noindent{}\hspace*{2.80em}"is\_valid": 1,\par
\noindent{}\hspace*{2.80em}"reason": "The tail entity is a country, which can be an item."\par
\noindent{}\hspace*{2.10em}\}\par
\noindent{}\hspace*{1.40em}\},\par
\noindent{}\hspace*{1.40em}\{\par
\noindent{}\hspace*{2.10em}"relation\_name": "country of citizenship",\par
\noindent{}\hspace*{2.10em}"relation\_description": "The object is a country that recognizes the subject as its citizen.",\par
\noindent{}\hspace*{2.10em}"semantic\_evaluation": \{\par
\noindent{}\hspace*{2.80em}"is\_match": 0,\par
\noindent{}\hspace*{2.80em}"reason": "The text is not about citizenship."\par
\noindent{}\hspace*{2.10em}\},\par
\noindent{}\hspace*{2.10em}"head\_entity\_evaluation": \{\par
\noindent{}\hspace*{2.80em}"is\_valid": 0,\par
\noindent{}\hspace*{2.80em}"reason": "The head entity is an organization, not a person."\par
\noindent{}\hspace*{2.10em}\},\par
\noindent{}\hspace*{2.10em}"tail\_entity\_evaluation": \{\par
\noindent{}\hspace*{2.80em}"is\_valid": 1,\par
\noindent{}\hspace*{2.80em}"reason": "The tail entity is a country, which is a valid type for this relation's range."\par
\noindent{}\hspace*{2.10em}\}\par
\noindent{}\hspace*{1.40em}\},\par
\noindent{}\hspace*{1.40em}\{\par
\noindent{}\hspace*{2.10em}"relation\_name": "occupant",\par
\noindent{}\hspace*{2.10em}"relation\_description": "A person or organization occupying property.",\par
\noindent{}\hspace*{2.10em}"semantic\_evaluation": \{\par
\noindent{}\hspace*{2.80em}"is\_match": 0,\par
\noindent{}\hspace*{2.80em}"reason": "The text does not state that the Local Government Association occupies property called 'England and Wales'."\par
\noindent{}\hspace*{2.10em}\},\par
\noindent{}\hspace*{2.10em}"head\_entity\_evaluation": \{\par
\noindent{}\hspace*{2.80em}"is\_valid": 1,\par
\noindent{}\hspace*{2.80em}"reason": "The head entity is an organization, which can be an occupant."\par
\noindent{}\hspace*{2.10em}\},\par
\noindent{}\hspace*{2.10em}"tail\_entity\_evaluation": \{\par
\noindent{}\hspace*{2.80em}"is\_valid": 0,\par
\noindent{}\hspace*{2.80em}"reason": "The tail entity is a country, not property."\par
\noindent{}\hspace*{2.10em}\}\par
\noindent{}\hspace*{1.40em}\},\par
\noindent{}\hspace*{1.40em}\{\par
\noindent{}\hspace*{2.10em}"relation\_name": "location",\par
\noindent{}\hspace*{2.10em}"relation\_description": "The location of the item, physical object or event is within.",\par
\noindent{}\hspace*{2.10em}"semantic\_evaluation": \{\par
\noindent{}\hspace*{2.80em}"is\_match": 0,\par
\noindent{}\hspace*{2.80em}"reason": "The text describes the association's area of operation, not its physical location. The relation is about location, not jurisdiction or service area."\par
\noindent{}\hspace*{2.10em}\},\par
\noindent{}\hspace*{2.10em}"head\_entity\_evaluation": \{\par
\noindent{}\hspace*{2.80em}"is\_valid": 1,\par
\noindent{}\hspace*{2.80em}"reason": "The head entity is an organization, which can have a location."\par
\noindent{}\hspace*{2.10em}\},\par
\noindent{}\hspace*{2.10em}"tail\_entity\_evaluation": \{\par
\noindent{}\hspace*{2.80em}"is\_valid": 1,\par
\noindent{}\hspace*{2.80em}"reason": "The tail entity is a country, which is a location."\par
\noindent{}\hspace*{2.10em}\}\par
\noindent{}\hspace*{1.40em}\},\par
\noindent{}\hspace*{1.40em}\{\par
\noindent{}\hspace*{2.10em}"relation\_name": "tributary",\par
\noindent{}\hspace*{2.10em}"relation\_description": "A stream or river that flows into this main stem (or parent) river.",\par
\noindent{}\hspace*{2.10em}"semantic\_evaluation": \{\par
\noindent{}\hspace*{2.80em}"is\_match": 0,\par
\noindent{}\hspace*{2.80em}"reason": "The text is not about geography or rivers."\par
\noindent{}\hspace*{2.10em}\},\par
\noindent{}\hspace*{2.10em}"head\_entity\_evaluation": \{\par
\noindent{}\hspace*{2.80em}"is\_valid": 0,\par
\noindent{}\hspace*{2.80em}"reason": "The head entity is an organization, not a river."\par
\noindent{}\hspace*{2.10em}\},\par
\noindent{}\hspace*{2.10em}"tail\_entity\_evaluation": \{\par
\noindent{}\hspace*{2.80em}"is\_valid": 0,\par
\noindent{}\hspace*{2.80em}"reason": "The tail entity is a country, not a river."\par
\noindent{}\hspace*{2.10em}\}\par
\noindent{}\hspace*{1.40em}\},\par
\noindent{}\hspace*{1.40em}\{\par
\noindent{}\hspace*{2.10em}"relation\_name": "nominated for",\par
\noindent{}\hspace*{2.10em}"relation\_description": "Award nomination received by a person, organisation or creative work.",\par
\noindent{}\hspace*{2.10em}"semantic\_evaluation": \{\par
\noindent{}\hspace*{2.80em}"is\_match": 0,\par
\noindent{}\hspace*{2.80em}"reason": "The text does not mention any awards."\par
\noindent{}\hspace*{2.10em}\},\par
\noindent{}\hspace*{2.10em}"head\_entity\_evaluation": \{\par
\noindent{}\hspace*{2.80em}"is\_valid": 1,\par
\noindent{}\hspace*{2.80em}"reason": "The head entity is an organization, which can be nominated for an award."\par
\noindent{}\hspace*{2.10em}\},\par
\noindent{}\hspace*{2.10em}"tail\_entity\_evaluation": \{\par
\noindent{}\hspace*{2.80em}"is\_valid": 0,\par
\noindent{}\hspace*{2.80em}"reason": "The tail entity is a country, not an award."\par
\noindent{}\hspace*{2.10em}\}\par
\noindent{}\hspace*{1.40em}\},\par
\noindent{}\hspace*{1.40em}\{\par
\noindent{}\hspace*{2.10em}"relation\_name": "original language of film or TV show",\par
\noindent{}\hspace*{2.10em}"relation\_description": "The language in which a film or a performance work was originally created.",\par
\noindent{}\hspace*{2.10em}"semantic\_evaluation": \{\par
\noindent{}\hspace*{2.80em}"is\_match": 0,\par
\noindent{}\hspace*{2.80em}"reason": "The text is not about films or TV shows."\par
\noindent{}\hspace*{2.10em}\},\par
\noindent{}\hspace*{2.10em}"head\_entity\_evaluation": \{\par
\noindent{}\hspace*{2.80em}"is\_valid": 0,\par
\noindent{}\hspace*{2.80em}"reason": "The head entity is an organization, not a film or TV show."\par
\noindent{}\hspace*{2.10em}\},\par
\noindent{}\hspace*{2.10em}"tail\_entity\_evaluation": \{\par
\noindent{}\hspace*{2.80em}"is\_valid": 0,\par
\noindent{}\hspace*{2.80em}"reason": "The tail entity is a country, not a language."\par
\noindent{}\hspace*{2.10em}\}\par
\noindent{}\hspace*{1.40em}\},\par
\noindent{}\hspace*{1.40em}\{\par
\noindent{}\hspace*{2.10em}"relation\_name": "member of",\par
\noindent{}\hspace*{2.10em}"relation\_description": "The organization or club to which the subject belongs.",\par
\noindent{}\hspace*{2.10em}"semantic\_evaluation": \{\par
\noindent{}\hspace*{2.80em}"is\_match": 0,\par
\noindent{}\hspace*{2.80em}"reason": "The text does not state that the Local Government Association is a member of England and Wales."\par
\noindent{}\hspace*{2.10em}\},\par
\noindent{}\hspace*{2.10em}"head\_entity\_evaluation": \{\par
\noindent{}\hspace*{2.80em}"is\_valid": 1,\par
\noindent{}\hspace*{2.80em}"reason": "The head entity is an organization, which can be a member of another organization."\par
\noindent{}\hspace*{2.10em}\},\par
\noindent{}\hspace*{2.10em}"tail\_entity\_evaluation": \{\par
\noindent{}\hspace*{2.80em}"is\_valid": 0,\par
\noindent{}\hspace*{2.80em}"reason": "The tail entity is a country, not an organization."\par
\noindent{}\hspace*{2.10em}\}\par
\noindent{}\hspace*{1.40em}\},\par
\noindent{}\hspace*{1.40em}\{\par
\noindent{}\hspace*{2.10em}"relation\_name": "sports season of league or competition",\par
\noindent{}\hspace*{2.10em}"relation\_description": "The property that shows the competition of which the item is a season.",\par
\noindent{}\hspace*{2.10em}"semantic\_evaluation": \{\par
\noindent{}\hspace*{2.80em}"is\_match": 0,\par
\noindent{}\hspace*{2.80em}"reason": "The text is not about sports."\par
\noindent{}\hspace*{2.10em}\},\par
\noindent{}\hspace*{2.10em}"head\_entity\_evaluation": \{\par
\noindent{}\hspace*{2.80em}"is\_valid": 0,\par
\noindent{}\hspace*{2.80em}"reason": "The head entity is an organization, not a sports season."\par
\noindent{}\hspace*{2.10em}\},\par
\noindent{}\hspace*{2.10em}"tail\_entity\_evaluation": \{\par
\noindent{}\hspace*{2.80em}"is\_valid": 0,\par
\noindent{}\hspace*{2.80em}"reason": "The tail entity is a country, not a competition."\par
\noindent{}\hspace*{2.10em}\}\par
\noindent{}\hspace*{1.40em}\},\par
\noindent{}\hspace*{1.40em}\{\par
\noindent{}\hspace*{2.10em}"relation\_name": "instrument",\par
\noindent{}\hspace*{2.10em}"relation\_description": "The musical instrument that a person plays.",\par
\noindent{}\hspace*{2.10em}"semantic\_evaluation": \{\par
\noindent{}\hspace*{2.80em}"is\_match": 0,\par
\noindent{}\hspace*{2.80em}"reason": "The text is not about music."\par
\noindent{}\hspace*{2.10em}\},\par
\noindent{}\hspace*{2.10em}"head\_entity\_evaluation": \{\par
\noindent{}\hspace*{2.80em}"is\_valid": 0,\par
\noindent{}\hspace*{2.80em}"reason": "The head entity is an organization, not a person."\par
\noindent{}\hspace*{2.10em}\},\par
\noindent{}\hspace*{2.10em}"tail\_entity\_evaluation": \{\par
\noindent{}\hspace*{2.80em}"is\_valid": 0,\par
\noindent{}\hspace*{2.80em}"reason": "The tail entity is a country, not an instrument."\par
\noindent{}\hspace*{2.10em}\}\par
\noindent{}\hspace*{1.40em}\},\par
\noindent{}\hspace*{1.40em}\{\par
\noindent{}\hspace*{2.10em}"relation\_name": "competition class",\par
\noindent{}\hspace*{2.10em}"relation\_description": "The official classification by a regulating body under which the subject qualifies for inclusion.",\par
\noindent{}\hspace*{2.10em}"semantic\_evaluation": \{\par
\noindent{}\hspace*{2.80em}"is\_match": 0,\par
\noindent{}\hspace*{2.80em}"reason": "The text is not about sports or competitions."\par
\noindent{}\hspace*{2.10em}\},\par
\noindent{}\hspace*{2.10em}"head\_entity\_evaluation": \{\par
\noindent{}\hspace*{2.80em}"is\_valid": 0,\par
\noindent{}\hspace*{2.80em}"reason": "The head entity is an organization, not a competition or class."\par
\noindent{}\hspace*{2.10em}\},\par
\noindent{}\hspace*{2.10em}"tail\_entity\_evaluation": \{\par
\noindent{}\hspace*{2.80em}"is\_valid": 0,\par
\noindent{}\hspace*{2.80em}"reason": "The tail entity is a country, not a classification."\par
\noindent{}\hspace*{2.10em}\}\par
\noindent{}\hspace*{1.40em}\},\par
\noindent{}\hspace*{1.40em}\{\par
\noindent{}\hspace*{2.10em}"relation\_name": "occupation",\par
\noindent{}\hspace*{2.10em}"relation\_description": "The occupation of a person.",\par
\noindent{}\hspace*{2.10em}"semantic\_evaluation": \{\par
\noindent{}\hspace*{2.80em}"is\_match": 0,\par
\noindent{}\hspace*{2.80em}"reason": "The text is not about a person's occupation."\par
\noindent{}\hspace*{2.10em}\},\par
\noindent{}\hspace*{2.10em}"head\_entity\_evaluation": \{\par
\noindent{}\hspace*{2.80em}"is\_valid": 0,\par
\noindent{}\hspace*{2.80em}"reason": "The head entity is an organization, not a person."\par
\noindent{}\hspace*{2.10em}\},\par
\noindent{}\hspace*{2.10em}"tail\_entity\_evaluation": \{\par
\noindent{}\hspace*{2.80em}"is\_valid": 0,\par
\noindent{}\hspace*{2.80em}"reason": "The tail entity is a country, not an occupation."\par
\noindent{}\hspace*{2.10em}\}\par
\noindent{}\hspace*{1.40em}\},\par
\noindent{}\hspace*{1.40em}\{\par
\noindent{}\hspace*{2.10em}"relation\_name": "original network",\par
\noindent{}\hspace*{2.10em}"relation\_description": "The network(s) the radio or television show was originally aired on.",\par
\noindent{}\hspace*{2.10em}"semantic\_evaluation": \{\par
\noindent{}\hspace*{2.80em}"is\_match": 0,\par
\noindent{}\hspace*{2.80em}"reason": "The text is not about TV or radio."\par
\noindent{}\hspace*{2.10em}\},\par
\noindent{}\hspace*{2.10em}"head\_entity\_evaluation": \{\par
\noindent{}\hspace*{2.80em}"is\_valid": 0,\par
\noindent{}\hspace*{2.80em}"reason": "The head entity is an organization, not a TV show."\par
\noindent{}\hspace*{2.10em}\},\par
\noindent{}\hspace*{2.10em}"tail\_entity\_evaluation": \{\par
\noindent{}\hspace*{2.80em}"is\_valid": 0,\par
\noindent{}\hspace*{2.80em}"reason": "The tail entity is a country, not a network."\par
\noindent{}\hspace*{2.10em}\}\par
\noindent{}\hspace*{1.40em}\},\par
\noindent{}\hspace*{1.40em}\{\par
\noindent{}\hspace*{2.10em}"relation\_name": "mountain range",\par
\noindent{}\hspace*{2.10em}"relation\_description": "The range or subrange to which the geographical item belongs.",\par
\noindent{}\hspace*{2.10em}"semantic\_evaluation": \{\par
\noindent{}\hspace*{2.80em}"is\_match": 0,\par
\noindent{}\hspace*{2.80em}"reason": "The text is not about geography or mountain ranges."\par
\noindent{}\hspace*{2.10em}\},\par
\noindent{}\hspace*{2.10em}"head\_entity\_evaluation": \{\par
\noindent{}\hspace*{2.80em}"is\_valid": 0,\par
\noindent{}\hspace*{2.80em}"reason": "The head entity is an organization, not a geographical feature."\par
\noindent{}\hspace*{2.10em}\},\par
\noindent{}\hspace*{2.10em}"tail\_entity\_evaluation": \{\par
\noindent{}\hspace*{2.80em}"is\_valid": 0,\par
\noindent{}\hspace*{2.80em}"reason": "The tail entity is a country, not a mountain range."\par
\noindent{}\hspace*{2.10em}\}\par
\noindent{}\hspace*{1.40em}\},\par
\noindent{}\hspace*{1.40em}\{\par
\noindent{}\hspace*{2.10em}"relation\_name": "crosses",\par
\noindent{}\hspace*{2.10em}"relation\_description": "The obstacle (body of water, road, ...) which this bridge crosses over or this tunnel goes under.",\par
\noindent{}\hspace*{2.10em}"semantic\_evaluation": \{\par
\noindent{}\hspace*{2.80em}"is\_match": 0,\par
\noindent{}\hspace*{2.80em}"reason": "The text is not about bridges or tunnels."\par
\noindent{}\hspace*{2.10em}\},\par
\noindent{}\hspace*{2.10em}"head\_entity\_evaluation": \{\par
\noindent{}\hspace*{2.80em}"is\_valid": 0,\par
\noindent{}\hspace*{2.80em}"reason": "The head entity is an organization, not a bridge or tunnel."\par
\noindent{}\hspace*{2.10em}\},\par
\noindent{}\hspace*{2.10em}"tail\_entity\_evaluation": \{\par
\noindent{}\hspace*{2.80em}"is\_valid": 0,\par
\noindent{}\hspace*{2.80em}"reason": "The tail entity is a country, not an obstacle."\par
\noindent{}\hspace*{2.10em}\}\par
\noindent{}\hspace*{1.40em}\},\par
\noindent{}\hspace*{1.40em}\{\par
\noindent{}\hspace*{2.10em}"relation\_name": "operating system",\par
\noindent{}\hspace*{2.10em}"relation\_description": "The operating system (OS) on which a software works or the OS installed on hardware.",\par
\noindent{}\hspace*{2.10em}"semantic\_evaluation": \{\par
\noindent{}\hspace*{2.80em}"is\_match": 0,\par
\noindent{}\hspace*{2.80em}"reason": "The text is not about software or operating systems."\par
\noindent{}\hspace*{2.10em}\},\par
\noindent{}\hspace*{2.10em}"head\_entity\_evaluation": \{\par
\noindent{}\hspace*{2.80em}"is\_valid": 0,\par
\noindent{}\hspace*{2.80em}"reason": "The head entity is an organization, not software or hardware."\par
\noindent{}\hspace*{2.10em}\},\par
\noindent{}\hspace*{2.10em}"tail\_entity\_evaluation": \{\par
\noindent{}\hspace*{2.80em}"is\_valid": 0,\par
\noindent{}\hspace*{2.80em}"reason": "The tail entity is a country, not an operating system."\par
\noindent{}\hspace*{2.10em}\}\par
\noindent{}\hspace*{1.40em}\},\par
\noindent{}\hspace*{1.40em}\{\par
\noindent{}\hspace*{2.10em}"relation\_name": "constellation",\par
\noindent{}\hspace*{2.10em}"relation\_description": "The area of the celestial sphere of which the subject is a part.",\par
\noindent{}\hspace*{2.10em}"semantic\_evaluation": \{\par
\noindent{}\hspace*{2.80em}"is\_match": 0,\par
\noindent{}\hspace*{2.80em}"reason": "The text is not about astronomy."\par
\noindent{}\hspace*{2.10em}\},\par
\noindent{}\hspace*{2.10em}"head\_entity\_evaluation": \{\par
\noindent{}\hspace*{2.80em}"is\_valid": 0,\par
\noindent{}\hspace*{2.80em}"reason": "The head entity is an organization, not a celestial object."\par
\noindent{}\hspace*{2.10em}\},\par
\noindent{}\hspace*{2.10em}"tail\_entity\_evaluation": \{\par
\noindent{}\hspace*{2.80em}"is\_valid": 0,\par
\noindent{}\hspace*{2.80em}"reason": "The tail entity is a country, not a constellation."\par
\noindent{}\hspace*{2.10em}\}\par
\noindent{}\hspace*{1.40em}\},\par
\noindent{}\hspace*{1.40em}\{\par
\noindent{}\hspace*{2.10em}"relation\_name": "record label",\par
\noindent{}\hspace*{2.10em}"relation\_description": "The brand and trademark associated with the marketing of subject music recordings and music videos.",\par
\noindent{}\hspace*{2.10em}"semantic\_evaluation": \{\par
\noindent{}\hspace*{2.80em}"is\_match": 0,\par
\noindent{}\hspace*{2.80em}"reason": "The text is not about music or record labels."\par
\noindent{}\hspace*{2.10em}\},\par
\noindent{}\hspace*{2.10em}"head\_entity\_evaluation": \{\par
\noindent{}\hspace*{2.80em}"is\_valid": 0,\par
\noindent{}\hspace*{2.80em}"reason": "The head entity is an organization, not a music recording."\par
\noindent{}\hspace*{2.10em}\},\par
\noindent{}\hspace*{2.10em}"tail\_entity\_evaluation": \{\par
\noindent{}\hspace*{2.80em}"is\_valid": 0,\par
\noindent{}\hspace*{2.80em}"reason": "The tail entity is a country, not a record label."\par
\noindent{}\hspace*{2.10em}\}\par
\noindent{}\hspace*{1.40em}\},\par
\noindent{}\hspace*{1.40em}\{\par
\noindent{}\hspace*{2.10em}"relation\_name": "place served by transport hub",\par
\noindent{}\hspace*{2.10em}"relation\_description": "The territorial entity or entities served by this transport hub (airport, train station, etc.).",\par
\noindent{}\hspace*{2.10em}"semantic\_evaluation": \{\par
\noindent{}\hspace*{2.80em}"is\_match": 0,\par
\noindent{}\hspace*{2.80em}"reason": "The text is not about transport hubs."\par
\noindent{}\hspace*{2.10em}\},\par
\noindent{}\hspace*{2.10em}"head\_entity\_evaluation": \{\par
\noindent{}\hspace*{2.80em}"is\_valid": 0,\par
\noindent{}\hspace*{2.80em}"reason": "The head entity is an organization, not a transport hub."\par
\noindent{}\hspace*{2.10em}\},\par
\noindent{}\hspace*{2.10em}"tail\_entity\_evaluation": \{\par
\noindent{}\hspace*{2.80em}"is\_valid": 1,\par
\noindent{}\hspace*{2.80em}"reason": "The tail entity is a country, which is a territorial entity."\par
\noindent{}\hspace*{2.10em}\}\par
\noindent{}\hspace*{1.40em}\},\par
\noindent{}\hspace*{1.40em}\{\par
\noindent{}\hspace*{2.10em}"relation\_name": "subsidiary",\par
\noindent{}\hspace*{2.10em}"relation\_description": "The subsidiary of a company or organization.",\par
\noindent{}\hspace*{2.10em}"semantic\_evaluation": \{\par
\noindent{}\hspace*{2.80em}"is\_match": 0,\par
\noindent{}\hspace*{2.80em}"reason": "The text does not state that England and Wales is a subsidiary of the Local Government Association."\par
\noindent{}\hspace*{2.10em}\},\par
\noindent{}\hspace*{2.10em}"head\_entity\_evaluation": \{\par
\noindent{}\hspace*{2.80em}"is\_valid": 1,\par
\noindent{}\hspace*{2.80em}"reason": "The head entity is an organization, which can have subsidiaries."\par
\noindent{}\hspace*{2.10em}\},\par
\noindent{}\hspace*{2.10em}"tail\_entity\_evaluation": \{\par
\noindent{}\hspace*{2.80em}"is\_valid": 0,\par
\noindent{}\hspace*{2.80em}"reason": "The tail entity is a country, not a subsidiary organization."\par
\noindent{}\hspace*{2.10em}\}\par
\noindent{}\hspace*{1.40em}\}\par
\noindent{}\hspace*{0.70em}]\par
\noindent{}\}\par
\noindent{}</think>\par
\par
\noindent{}<summary>\par
\noindent{}residence,0,0,1\par
\noindent{}headquarters location,0,1,1\par
\noindent{}military rank,0,0,0\par
\noindent{}child,0,0,0\par
\noindent{}operator,0,1,0\par
\noindent{}heritage designation,0,0,0\par
\noindent{}field of work,0,1,0\par
\noindent{}position played on team / speciality,0,0,0\par
\noindent{}characters,0,0,0\par
\noindent{}applies to jurisdiction,1,1,1\par
\noindent{}sibling,0,0,0\par
\noindent{}sport,0,0,0\par
\noindent{}member of political party,0,0,0\par
\noindent{}position held,0,1,0\par
\noindent{}notable work,0,1,0\par
\noindent{}voice type,0,0,0\par
\noindent{}screenwriter,0,0,0\par
\noindent{}contains administrative territorial entity,0,0,1\par
\noindent{}location of formation,0,1,1\par
\noindent{}located in the administrative territorial entity,0,1,1\par
\noindent{}said to be the same as,0,1,1\par
\noindent{}country of citizenship,0,0,1\par
\noindent{}occupant,0,1,0\par
\noindent{}location,0,1,1\par
\noindent{}tributary,0,0,0\par
\noindent{}nominated for,0,1,0\par
\noindent{}original language of film or TV show,0,0,0\par
\noindent{}member of,0,1,0\par
\noindent{}sports season of league or competition,0,0,0\par
\noindent{}instrument,0,0,0\par
\noindent{}competition class,0,0,0\par
\noindent{}occupation,0,0,0\par
\noindent{}original network,0,0,0\par
\noindent{}mountain range,0,0,0\par
\noindent{}crosses,0,0,0\par
\noindent{}operating system,0,0,0\par
\noindent{}constellation,0,0,0\par
\noindent{}record label,0,0,0\par
\noindent{}place served by transport hub,0,0,1\par
\noindent{}subsidiary,0,1,0</summary>\par
\par
\noindent{}----------------------------------------\par
\noindent{}Parsed Result:\par
\noindent{}----------------------------------------\par
\noindent{}A layer (1 relations): ['applies to jurisdiction']\par
\noindent{}B layer (5 relations): ['headquarters location', 'location of formation', 'located in the administrative territorial entity', 'said to be the same as', 'location']\par
\noindent{}C layer (34 relations): ['residence', 'military rank', 'child', 'operator', 'heritage designation', 'field of work', 'position played on team / speciality', 'characters', 'sibling', 'sport', 'member of political party', 'position held', 'notable work', 'voice type', 'screenwriter', 'contains administrative territorial entity', 'country of citizenship', 'occupant', 'tributary', 'nominated for', 'original language of film or TV show', 'member of', 'sports season of league or competition', 'instrument', 'competition class', 'occupation', 'original network', 'mountain range', 'crosses', 'operating system', 'constellation', 'record label', 'place served by transport hub', 'subsidiary']\par
\par
\par
\endgroup

\subsection{Fine-Grained Comparative Reasoning Trace}
Given the fine-grained comparative reasoning prompt, $\mathcal{M}_{3}$ produces the following reasoning trace.
This trace shows how ReaORE makes the final relation prediction by comparing relation pairs rather than directly generating a label.
Each comparison records the pair, the preferred relation, and the reason for the preference.
The final decision summary then aggregates these pairwise judgments into one relation prediction, explaining why the selected relation is more appropriate than the alternatives.

\begingroup
\normalsize
\sloppy
\raggedright
\setlength{\parindent}{0pt}
\setlength{\parskip}{0pt}
\setlength{\leftskip}{0.75em}
\par
\noindent{}Query:\par
\noindent{}Input: Text: It is an affiliated member of the Local Government Association of England and Wales and the regional employers organisation .\par
\noindent{}Head entity: local government association.\par
\noindent{}Tail entity: england and wales.\par
\noindent{}Candidate: ['applies to jurisdiction', 'located in the administrative territorial entity', 'location', 'contains administrative territorial entity']\par
\noindent{}Output: <think>\par
\noindent{}\{\par
\noindent{}\hspace*{0.70em}"comparison\_log": [\par
\noindent{}\hspace*{1.40em}\{\par
\noindent{}\hspace*{2.10em}"pair": "applies to jurisdiction vs located in the administrative territorial entity",\par
\noindent{}\hspace*{2.10em}"winner": "applies to jurisdiction",\par
\noindent{}\hspace*{2.10em}"reason": "The text states the Local Government Association is 'of England and Wales', which means it operates within that jurisdiction. 'applies to jurisdiction' correctly captures this relationship. 'located in the administrative territorial entity' is incorrect because a national organization is not located within a specific country; it operates within its jurisdiction."\par
\noindent{}\hspace*{1.40em}\},\par
\noindent{}\hspace*{1.40em}\{\par
\noindent{}\hspace*{2.10em}"pair": "applies to jurisdiction vs location",\par
\noindent{}\hspace*{2.10em}"winner": "applies to jurisdiction",\par
\noindent{}\hspace*{2.10em}"reason": "While the association is in 'England and Wales', 'applies to jurisdiction' is a more precise and formal description of the relationship between an organization and the country it serves. 'location' is too generic and less precise for this context."\par
\noindent{}\hspace*{1.40em}\},\par
\noindent{}\hspace*{1.40em}\{\par
\noindent{}\hspace*{2.10em}"pair": "applies to jurisdiction vs contains administrative territorial entity",\par
\noindent{}\hspace*{2.10em}"winner": "applies to jurisdiction",\par
\noindent{}\hspace*{2.10em}"reason": "The relation 'contains administrative territorial entity' would imply that the Local Government Association is a larger entity that contains the country of England and Wales, which is factually incorrect. 'applies to jurisdiction' correctly describes the relationship."\par
\noindent{}\hspace*{1.40em}\},\par
\noindent{}\hspace*{1.40em}\{\par
\noindent{}\hspace*{2.10em}"pair": "located in the administrative territorial entity vs location",\par
\noindent{}\hspace*{2.10em}"winner": "located in the administrative territorial entity",\par
\noindent{}\hspace*{2.10em}"reason": "Both relations are poor fits. However, 'located in the administrative territorial entity' is slightly better than 'location' because it specifies the type of entity (administrative territorial), which is relevant to the context of a country like England and Wales. 'location' is too generic."\par
\noindent{}\hspace*{1.40em}\},\par
\noindent{}\hspace*{1.40em}\{\par
\noindent{}\hspace*{2.10em}"pair": "located in the administrative territorial entity vs contains administrative territorial entity",\par
\noindent{}\hspace*{2.10em}"winner": "located in the administrative territorial entity",\par
\noindent{}\hspace*{2.10em}"reason": "'contains administrative territorial entity' is factually incorrect as the association does not contain the country. 'located in the administrative territorial entity' is a weak but plausible interpretation of the text, even if not the most precise."\par
\noindent{}\hspace*{1.40em}\},\par
\noindent{}\hspace*{1.40em}\{\par
\noindent{}\hspace*{2.10em}"pair": "location vs contains administrative territorial entity",\par
\noindent{}\hspace*{2.10em}"winner": "location",\par
\noindent{}\hspace*{2.10em}"reason": "'contains administrative territorial entity' is factually incorrect. 'location' is a very general but not entirely wrong description of the relationship, as the association is situated in the countries of England and Wales."\par
\noindent{}\hspace*{1.40em}\}\par
\noindent{}\hspace*{0.70em}],\par
\noindent{}\hspace*{0.70em}"final\_result": \{\par
\noindent{}\hspace*{1.40em}"best\_relation": "applies to jurisdiction",\par
\noindent{}\hspace*{1.40em}"decision\_logic": "The text 'of England and Wales' indicates that the Local Government Association operates within the jurisdiction of England and Wales. This makes 'applies to jurisdiction' the most accurate and specific relation. The other candidates are either too generic ('location'), incorrectly describe the relationship ('contains administrative territorial entity'), or are factually incorrect ('located in the administrative territorial entity')."\par
\noindent{}\hspace*{0.70em}\}\par
\noindent{}\}\par
\noindent{}</think>\par
\par
\noindent{}\{\par
\noindent{}\hspace*{0.70em}"comparison\_log": [\par
\noindent{}\hspace*{1.40em}\{\par
\noindent{}\hspace*{2.10em}"pair": "applies to jurisdiction vs located in the administrative territorial entity",\par
\noindent{}\hspace*{2.10em}"winner": "applies to jurisdiction",\par
\noindent{}\hspace*{2.10em}"reason": "The phrase 'of England and Wales' indicates the association operates within that jurisdiction. 'applies to jurisdiction' is the correct relation. 'located in the administrative territorial entity' is incorrect because a national organization is not located within a country; it operates within its jurisdiction."\par
\noindent{}\hspace*{1.40em}\},\par
\noindent{}\hspace*{1.40em}\{\par
\noindent{}\hspace*{2.10em}"pair": "applies to jurisdiction vs location",\par
\noindent{}\hspace*{2.10em}"winner": "applies to jurisdiction",\par
\noindent{}\hspace*{2.10em}"reason": "'applies to jurisdiction' is more precise than the generic 'location'. The text implies a functional relationship (operating within a jurisdiction), not just a physical location."\par
\noindent{}\hspace*{1.40em}\},\par
\noindent{}\hspace*{1.40em}\{\par
\noindent{}\hspace*{2.10em}"pair": "applies to jurisdiction vs contains administrative territorial entity",\par
\noindent{}\hspace*{2.10em}"winner": "applies to jurisdiction",\par
\noindent{}\hspace*{2.10em}"reason": "'contains administrative territorial entity' is factually incorrect. The association does not contain the country of England and Wales. 'applies to jurisdiction' correctly describes the relationship."\par
\noindent{}\hspace*{1.40em}\},\par
\noindent{}\hspace*{1.40em}\{\par
\noindent{}\hspace*{2.10em}"pair": "located in the administrative territorial entity vs location",\par
\noindent{}\hspace*{2.10em}"winner": "located in the administrative territorial entity",\par
\noindent{}\hspace*{2.10em}"reason": "Both are weak fits. However, 'located in the administrative territorial entity' is slightly better because it specifies the type of entity (administrative territorial), which is relevant to the context of a country like England and Wales. 'location' is too generic."\par
\noindent{}\hspace*{1.40em}\},\par
\noindent{}\hspace*{1.40em}\{\par
\noindent{}\hspace*{2.10em}"pair": "located in the administrative territorial entity vs contains administrative territorial entity",\par
\noindent{}\hspace*{2.10em}"winner": "located in the administrative territorial entity",\par
\noindent{}\hspace*{2.10em}"reason": "'contains administrative territorial entity' is factually incorrect. 'located in the administrative territorial entity' is a weak but plausible interpretation of the text, even if not the most precise."\par
\noindent{}\hspace*{1.40em}\},\par
\noindent{}\hspace*{1.40em}\{\par
\noindent{}\hspace*{2.10em}"pair": "location vs contains administrative territorial entity",\par
\noindent{}\hspace*{2.10em}"winner": "location",\par
\noindent{}\hspace*{2.10em}"reason": "'contains administrative territorial entity' is factually incorrect. 'location' is a very general but not entirely wrong description of the relationship, as the association is situated in the countries of England and Wales."\par
\noindent{}\hspace*{1.40em}\}\par
\noindent{}\hspace*{0.70em}],\par
\noindent{}\hspace*{0.70em}"final\_result": \{\par
\noindent{}\hspace*{1.40em}"best\_relation": "applies to jurisdiction",\par
\noindent{}\hspace*{1.40em}"decision\_logic": "The text 'of England and Wales' indicates that the Local Government Association operates within the jurisdiction of England and Wales. This makes 'applies to jurisdiction' the most accurate and specific relation. The other candidates are either too generic ('location'), incorrectly describe the relationship ('contains administrative territorial entity'), or are factually incorrect ('located in the administrative territorial entity')."\par
\noindent{}\hspace*{0.70em}\}\par
\noindent{}\}\par
\noindent{}```\par
\noindent{}Parsed: \{'comparison\_log': [\{'pair': 'applies to jurisdiction vs located in the administrative territorial entity', 'winner': 'applies to jurisdiction', 'reason': "The text states the Local Government Association is 'of England and Wales', which means it operates within that jurisdiction. 'applies to jurisdiction' correctly captures this relationship. 'located in the administrative territorial entity' is incorrect because a national organization is not located within a specific country; it operates within its jurisdiction."\}, \{'pair': 'applies to jurisdiction vs location', 'winner': 'applies to jurisdiction', 'reason': "While the association is in 'England and Wales', 'applies to jurisdiction' is a more precise and formal description of the relationship between an organization and the country it serves. 'location' is too generic and less precise for this context."\}, \{'pair': 'applies to jurisdiction vs contains administrative territorial entity', 'winner': 'applies to jurisdiction', 'reason': "The relation 'contains administrative territorial entity' would imply that the Local Government Association is a larger entity that contains the country of England and Wales, which is factually incorrect. 'applies to jurisdiction' correctly describes the relationship."\}, \{'pair': 'located in the administrative territorial entity vs location', 'winner': 'located in the administrative territorial entity', 'reason': "Both relations are poor fits. However, 'located in the administrative territorial entity' is slightly better than 'location' because it specifies the type of entity (administrative territorial), which is relevant to the context of a country like England and Wales. 'location' is too generic."\}, \{'pair': 'located in the administrative territorial entity vs contains administrative territorial entity', 'winner': 'located in the administrative territorial entity', 'reason': "'contains administrative territorial entity' is factually incorrect as the association does not contain the country. 'located in the administrative territorial entity' is a weak but plausible interpretation of the text, even if not the most precise."\}, \{'pair': 'location vs contains administrative territorial entity', 'winner': 'location', 'reason': "'contains administrative territorial entity' is factually incorrect. 'location' is a very general but not entirely wrong description of the relationship, as the association is situated in the countries of England and Wales."\}], 'final\_result': \{'best\_relation': 'applies to jurisdiction', 'decision\_logic': "The text 'of England and Wales' indicates that the Local Government Association operates within the jurisdiction of England and Wales. This makes 'applies to jurisdiction' the most accurate and specific relation. The other candidates are either too generic ('location'), incorrectly describe the relationship ('contains administrative territorial entity'), or are factually incorrect ('located in the administrative territorial entity')."\}\}\par
\par
\par
\endgroup

\end{document}